\documentclass[10pt,journal,compsoc]{IEEEtran}

\ifCLASSOPTIONcompsoc
  \usepackage[nocompress]{cite}
\else
  \usepackage{cite}
\fi

\ifCLASSINFOpdf
\else
\fi

\usepackage{epsfig}
\usepackage{helvet}
\usepackage{courier}
\usepackage{floatrow}
\usepackage{multirow}
\newfloatcommand{capbtabbox}{table}[][\FBwidth]
\usepackage{comment}
\usepackage[utf8]{inputenc}
\usepackage[T1]{fontenc}
\usepackage{amsfonts}
\usepackage{units}
\usepackage{microtype}
\usepackage[dvipsnames]{xcolor}
\usepackage{soul}
\usepackage{url}
\usepackage[small]{caption}
\usepackage{graphicx}
\usepackage{amsmath}
\usepackage{booktabs}
\usepackage{epstopdf}
\usepackage{amsthm}
\usepackage{amssymb}
\usepackage{subfig}
\usepackage{wrapfig}

\usepackage{nicefrac}
\usepackage{caption}
\usepackage{amssymb}
\usepackage{diagbox}
\usepackage{extpfeil}
\usepackage{makecell}
\usepackage{adjustbox}
\usepackage{color,colortbl}

\usepackage{bbding}
\usepackage{enumitem}
\usepackage{float}
\usepackage{multicol}
\usepackage[colorlinks,
    linkcolor=red,
    anchorcolor=blue,
    citecolor=green
]{hyperref}

\usepackage[ruled,vlined]{algorithm2e}
\usepackage{algpseudocode}

\hypersetup{
hidelinks,
colorlinks=true
}

\definecolor{lightgray}{gray}{0.95}
\definecolor{color3}{gray}{0.95}
\definecolor{rouse}{rgb}{0.981,0.961,0.941}
\definecolor{light-yellow}{rgb}{1,1,0.93}
\definecolor{light-green}{rgb}{0.95,1,0.95}

\definecolor{colorTabTop}{rgb}{0.93,0.92,0.94}
\definecolor{colorTab}{rgb}{0.92,0.95,0.92}

\usepackage{pgfplots}
\usepackage{pgfplotstable}
\usepackage{booktabs}
\usepackage{amsmath}
\usepgfplotslibrary{groupplots}

\pgfplotsset{
    compat=1.18,
    tick label style={font=\small},
    label style={font=\small},
    legend style={font=\scriptsize},
    title style={font=\small\bfseries}
}

\definecolor{set5}{RGB}{228,26,28}
\definecolor{set14}{RGB}{55,126,184}
\definecolor{b100}{RGB}{77,175,74}
\definecolor{urban100}{RGB}{152,78,163}
\definecolor{manga109}{RGB}{255,127,0}

\begin{document}
\title{QuantSR+: Pushing the Limit of Quantized Image Super-Resolution Networks}

\author{
Haotong~Qin,
Xudong~Ma,
Xianglong~Liu,~\IEEEmembership{Senior~Member,~IEEE},
Jie~Luo,
Jinyang~Guo,
Michele~Magno,~\IEEEmembership{Fellow,~IEEE},
Yulun~Zhang$^{ *}$
\IEEEcompsocitemizethanks{
\IEEEcompsocthanksitem
H. Qin and M. Magno are from ETH Zurich, Switzerland.
Y. Zhang ($^*$Corresponding Author, E-mail: yulun100@gmail.com) is from Shanghai Jiao Tong University, China.
X. Ma, X. Liu, J. Luo, and J. Guo are from Beihang University, China.
}
}

\markboth{Journal of \LaTeX\ Class Files,~Vol.~14, No.~8, August~2015}
{Shell \MakeLowercase{\textit{et al.}}: Bare Demo of IEEEtran.cls for Computer Society Journals}

\IEEEtitleabstractindextext{
\begin{abstract}
Low-bit quantization is widely used to compress super-resolution (SR) models and reduce storage and computation costs for deployment on resource-limited devices. However, when SR models are pushed to ultra-low precision (2-4 bits), performance can drop sharply due to diminished representational capacity and the detail-sensitive nature of SR.
To address these issues, we propose \textbf{QuantSR+}, a unified framework that improves quantization operators, network design, and training optimization, achieving better trade-offs between accuracy and efficiency than prior low-bit SR methods. QuantSR+ mainly relies on three technical contributions: (1) \textit{Redistribution-driven Bit Determination} (RBD), which reshapes quantization distributions in both forward and backward passes to preserve representation fidelity; (2) \textit{Quantized Slimmable Architecture} (QSA), which begins with an over-parameterized model and progressively prunes less critical blocks to meet efficiency budgets while pushing the accuracy performance; and (3) \textit{Slimming-guided Function-localized Distillation} (SFD), which enforces block-aware feature alignment via a direct loss and a progressive, function-local training schedule to capture quantization effects better and speed up convergence.
Extensive experiments show that QuantSR+ achieves state-of-the-art performance against both specialized quantized SR methods and generic quantization approaches. For SwinIR-S on Urban100 ($\times$4), it improves PSNR by 0.29 dB over the 2-bit SOTA baseline. Meanwhile, it delivers strong efficiency gains at 2-bit, reducing operations by up to 87.9\% and storage by 89.4\%. QuantSR+ is effective for both convolutional and transformer-based SR models, indicating broad applicability. Code and pretrained models are available \href{https://github.com/Efficient-ML/QuantSRv2}{here}.
\end{abstract}

}

\maketitle

\IEEEdisplaynontitleabstractindextext

\IEEEpeerreviewmaketitle

\begin{figure*}[t]
\centering
\includegraphics[width=\textwidth]{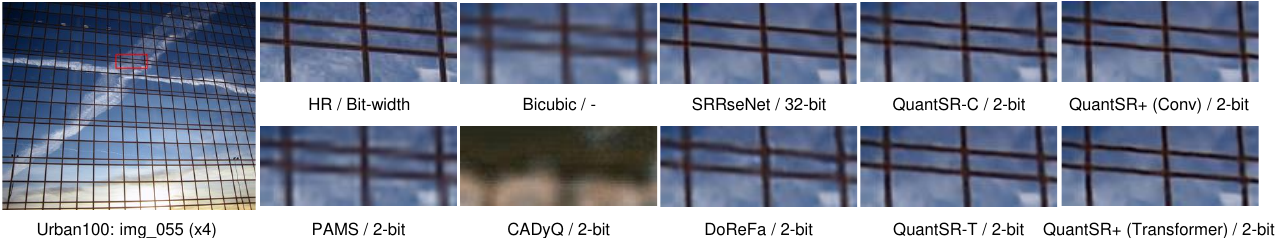} \\
\caption{Visualization for quantized SR models in Urban100 $\times$4 scale, and SRResNet~\cite{ledig2017photo} is used as the full-precision SR model. QuantSR+ outperforms recent quantization methods consistently, including DoReFa~\cite{zhou2016dorefa}, PAMS~\cite{li2020pams}, CADyQ~\cite{hong2022cadyq}, and QuantSR~\cite{qin2024quantsr}.}
\label{fig:quantsr_first_fig_SRBIX4_lightweight}
\end{figure*}

\IEEEraisesectionheading{\section{Introduction}}
\label{sec:intro}
\IEEEPARstart{W}{ith} \textcolor{black}{the rapid development of deep neural networks, the demand for image super-resolution (SR) applications is widespread across various scenarios~\cite{kim2016deeply,lim2017enhanced}. The SR task aims to reconstruct a high-resolution (HR) version from a low-resolution (LR) input by inferring high-frequency details, but it is inherently ill-posed because an LR image may correspond to multiple possible HR candidates~\cite{sun2024learning,liu2022blind,guo2024towards,su2022global,zhang2021residual}.
Existing efforts typically focus on convolution-based or transformer-based network designs to achieve high reconstruction fidelity~\cite{zhang2018densely,diffbir,yue2024arbitrary}, and recent diffusion-based SR further improves the perceptual quality of SR reconstruction~\cite{wang2024exploiting,wu2024one}. Due to the large volume of parameters and high computational complexity, current image SR models often demand costly hardware resources for real-time deployment, posing challenges in resource-constrained scenarios such as mobile or edge devices.
Therefore, model compression approaches are urgently needed to achieve lightweight models and efficient inference. Among various compression approaches, low-bit quantization effectively compresses and accelerates models by reducing parameter bit-width: by quantizing computation-intensive units such as convolutions, matrix multiplications, and linear layers~\cite{choukroun2019omse,hubara2021accurate,tu2023toward,hong2024overcoming,sun20242}, quantized models can use efficient integer operations instead of floating-point computations, which is particularly important for deploying advanced artificial intelligence applications on edge devices; moreover, quantization mainly compresses parameters within layers on an element-wise basis, making it widely applicable across architectures and tasks, and the quantization function (quantizer) can be customized to specific bit-widths, typically 2 to 8 bits, to flexibly balance accuracy and efficiency~\cite{tian2023cabm,zhang2024anycost,hong2024adabm,makhov2024towards,wang2025outlier,wang2025thinking}.}

Although low-bit weights and activations can enhance computational efficiency, quantization may significantly degrade or even destabilize performance in image SR tasks, especially when employing ultra-low bit-widths, \textit{e.g.}, 2-4 bits. While existing studies have attempted to minimize quantization losses, a considerable performance gap remains between low-bit SR models and their full-precision counterparts. We attribute this performance degradation primarily to the following three factors:
(1) \textbf{Operator Degradation}: Particularly in ultra-low bit-width quantization, the bit-wise discretization process results in highly degraded and inaccurate representations within the computation units of SR models, limiting the quality and diversity of the output of the SR model.
(2) \textbf{Structure Constraints}: Since quantization occurs at the parameter level, the accuracy of the full-precision model with the same structure can almost be seen as the upper limit for the quantized versions, making it challenging to reach an optimal solution in the discrete parameter space.
(3) \textbf{Optimization Perturbations}: The quantization errors introduced during discretization are difficult to estimate directly and tend to accumulate during training, thereby complicating the stable convergence of quantized SR models.

To address these challenges encountered in low-bit quantization, we propose an accurate, efficient, and flexible \textbf{Quant}ized image \textbf{S}uper-\textbf{R}esolution model, \textbf{QuantSR+}, designed to push the performance of SR models under low bit-widths to the limit. At the operator level, we introduce \textit{Redistribution-driven Bit Determination} (RBD), which optimizes and reconstructs bit-level quantization function representations during both forward and backward propagation through propagation-based redistribution. Compared to vanilla quantization, RBD enhances the accuracy of quantized units within the SR model without increasing the inference overhead through bit-level reorganization. At the structure level, we develop the \textit{Quantized Slimmable Architecture} (QSA), which starts from an over-parameterized initial structure that is two times larger than the targeted compressed model. It aims to push the upper accuracy limits of compressed models. It allows for progressive structural evolution during training, intending to surpass the accuracy set by the original counterparts under specific inference efficiency constraints.
At the optimization level, we implement \textit{Slimming-guided Function-localized Distillation} (SFD), which introduces a direct loss to capture quantization effects and a progressive strategy for optimizing network blocks based on their local functions. By adjusting the granularity of internal feature alignment, SFD effectively supervises the quantized SR model, ensuring both stability and faster convergence.

We conduct extensive experiments to demonstrate that the performance of QuantSR+ significantly surpasses various bit-width quantized SR models (see Fig.~\ref{fig:quantsr_first_fig_SRBIX4_lightweight}), with 2-bit QuantSR+ performing comparably to existing state-of-the-art methods using 4 bits. We also verified the versatility of QuantSR+ on convolutional and transformer-based SR networks, along with its theoretical and deployment efficiency, showcasing its adaptability across different tasks, architectures, and scenarios. Comprehensive quantization and visualization results show that QuantSR+ surpasses current state-of-the-art quantized SR networks and general quantization approaches, achieving a significant improvement of 0.29 dB in PSNR over 2-bit existing SOTA methods on the $\times$4 Urban100 dataset. QuantSR+ demonstrates comprehensive accuracy and efficiency enhancements across both convolutional and transformer-based architectures, validating the versatility of the proposed techniques in various architectures. QuantSR+ can also achieve exceptional computational efficiency. We report the inference efficiency of 4-bit and 2-bit QuantSR+, and the 2-bit version even achieves an 87.9\% reduction of computational FLOPs and 88.2\% reduction of storage usage compared to its full-precision counterpart.

Note that the paper is an extension of the original conference version~\cite{qin2024quantsr}. Compared with the conference version, the main contributions of the paper are

\begin{itemize}

\item  This paper further identifies the causes of significant performance degradation of quantized SR models at ultra-low bit-widths, comprehensively pointing out the issues from operator degradation, structure constraints, and optimization perturbations as the primary challenges faced by quantized SR models. This provides empirical inspiration for QuantSR+.

\item  \textcolor{black}{We propose QuantSR+, an accurate and efficient algorithm designed to push the performance of low-bit super-resolution models to their limits. In particular, we emphasize that QuantSR+ introduces comprehensive improvements over the original conference paper in both the quantizer design and the model architecture, and further presents a new distillation-based optimization strategy. As a result, compared with the method in the original conference version, QuantSR+ offers stronger representational capacity and is easier to optimize:}

\noindent \textcolor{black}{\textit{(1)} We improve the quantizer via bit-determination redistribution.
Compared with the conference paper, this technique refines redistribution in the weight quantizer down to the individual-bit level, improving the flexibility and accuracy of the quantized representation. More importantly, the packed quantized weights produced by the quantizer are exactly consistent with standard integers and so incur no additional inference overhead.}

\noindent \textcolor{black}{\textit{(2)} We improve the quantized architecture to break through its accuracy limits.
Unlike the conference paper, which abruptly changes the entire architecture at a certain point, this technique distributes the slimming process across blocks and stages of training. It better couples architectural evolution with the training procedure, enabling the quantized model to be continuously optimized throughout training.}

\noindent \textcolor{black}{\textit{(3)} We propose a novel function-localized distillation for quantization-aware training.
Compared with the standard training pipeline used in the conference version, this technique introduces a distillation loss based on localized functional reconstruction, allowing targeted optimization of the slimmed quantized architecture. It helps the quantized model remain stable during slimming evolution and converge faster.}

\item  Further detailed evaluations and analyses of the quantized SR networks are presented, including more quantitative and visual results for image SR tasks and statistical analysis of parameters during training, showcasing the significant accuracy and efficiency improvements brought by QuantSR+. We also report the computational Ops and storage savings of the proposed QuantSR+, demonstrating its efficiency potential in resource-constrained scenarios. The results show that significant accuracy improvements are achieved without additional efficiency, which strongly demonstrates the advantages of QuantSR+ over existing methods.

\end{itemize}

\section{Related Work}

\subsubsection{Efficient Image SR Model}

Convolutional neural networks have consistently exhibited remarkable performance in image SR. EDSR~\cite{lim2017enhanced} and SRResNet~\cite{ledig2017photo} already outperformed traditional methods and have since served as a pivotal CNN-based reference. Vision Transformers (ViT)~\cite{dosovitskiy2020image} have catalyzed the emergence of Transformer-driven approaches, exemplified by SwinIR~\cite{liang2021swinir}, which surpass many CNN models. Meanwhile, efficiency-oriented strategies such as Neural Architecture Search (NAS)~\cite{Chen2020a}, Knowledge Distillation (KD)~\cite{hinton2014distilling}, and pruning~\cite{OBS} have been broadly investigated to address resource constraints. Efforts include multi-distillation networks to mitigate redundancy \cite{hui2019lightweight}, cascading residual networks for enhanced efficiency \cite{ahn2018fast}, and NAS-based approaches like FALSR \cite{chu2019fast}. KD has been adopted to train lightweight SR models \cite{he2020fakd,lee2020learning}, while pruning, as in the ASSL framework \cite{zhang2021aligned}, systematically removes surplus parameters. Although these methods are effective, they can overlook subtle redundancies of computation and storage caused by the expensive floating-point parameters in networks. Consequently, low-bit quantization stands out as a promising compression approach for shrinking bit-width and computational costs while sustaining performance in real-world SR applications.

\subsubsection{Model Quantization}

\textcolor{black}{Among the various quantization techniques, Quantization-Aware Training (QAT)~\cite{zhou2016dorefa,tian2023cabm,hong2024adabm,wang2025thinking} and Post-Training Quantization (PTQ)~\cite{hubara2021accurate,tu2023toward,wang2025outlier} have emerged as the primary approaches for reducing computational and memory overhead in deep neural networks~\cite{Gholami2022,Choi2018a}.}
QAT incorporates quantization directly into the training process, thus leveraging rich training signals to support aggressive low-bit (2-4 bits) quantization~\cite{zhou2016dorefa,choi2018pact,li2020pams,hong2022cadyq}. By fine-tuning parameters while accounting for quantization effects, QAT enables substantial efficiency gains while mitigating accuracy degradation.
In contrast, PTQ allows for quantizing pre-trained models without further training~\cite{hubara2021accurate}, offering a more straightforward and widely adopted approach~\cite{choukroun2019omse,jhunjhunwala2021adaquant,hubara2021accurate}. Despite its simplicity, PTQ tends to underperform at extremely low bit widths due to its reliance on fixed parameters and the absence of retraining or fine-tuning. These limitations become pronounced when models are compressed to ultra-low-bit configurations.

\subsubsection{Quantized SR Models}

For SR tasks, low-bit quantization presents an effective approach to address the computational bottlenecks associated with high-resolution input data. SR models notoriously demand considerable floating-point operations and storage, making them prime candidates for compression strategies that can alleviate resource constraints~\cite{wang2021fully,ignatov2022efficient,liu2021super}. \textcolor{black}{As a result, researchers began to explore QAT-based methods~\cite{tian2023cabm,hong2024adabm,wang2025thinking} to balance efficiency and accuracy in SR networks. Hong et al. proposed a distribution-aware quantization (DAQ) method that applies channel-wise, data-driven quantization parameters~\cite{hong2022daq}}, while CADyQ employs a strategic bit-allocation scheme across different layers and image regions~\cite{hong2022cadyq}. Similarly, PAMS leverages a trainable truncated parameter to dynamically adjust the upper bound of the quantization range, bolstering flexibility and performance~\cite{li2020pams}.
Despite these advances, a notable performance gap remains between fully quantized SR models and their full-precision counterparts. Bridging this gap continues to be a critical research focus, as fully harnessing the benefits of low-bit quantization is essential for practical SR applications on resource-constrained devices.

\section{Method}
This section introduces the proposed \textbf{QuantSR+} for accurate and efficient image SR tasks. Firstly, we present the existing baselines for quantizing SR models, including the architecture definition of SR models as well as the quantization and computation processes. Subsequently, we detail the novel techniques in QuantSR+ to address various limitations of the existing baselines, including \textit{Redistribution-driven Bit Determination} (RBD), \textit{Quantized Slimmable Architecture} (QSA), and \textit{Slimming-guided Function-localized Distillation} (SFD). Finally, we present how to apply QuantSR+ to image SR tasks and discuss the training pipeline and specific implementation details for quantized SR models.

\begin{figure*}[t]
\centering
\includegraphics[width=0.99\linewidth]{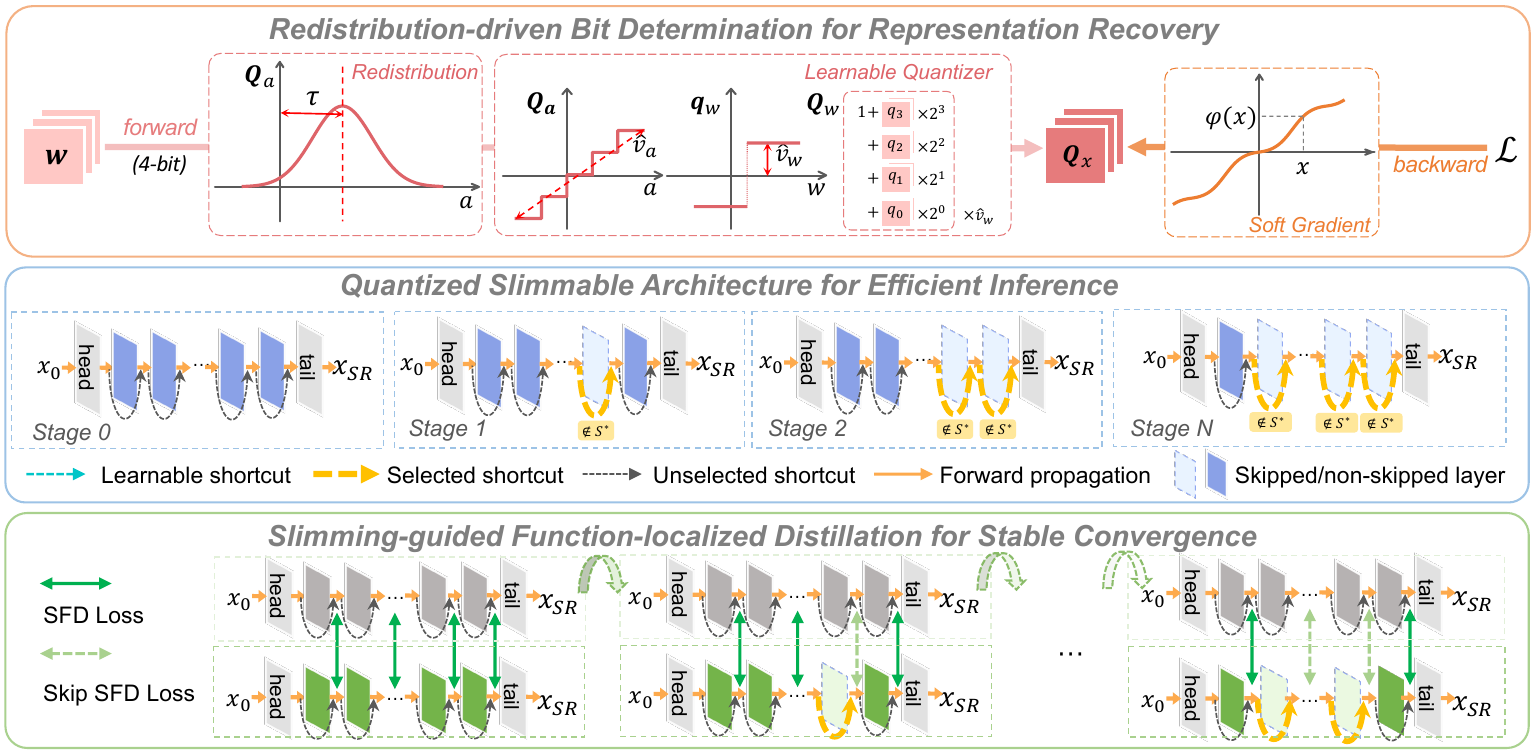}
\caption{
Overview of the proposed QuantSR+ for image SR. The QuantSR+ improves the quantized SR network at the operator (RBD), structure (QSA), and optimization (SFD) levels comprehensively.
}
\label{fig:overview}
\end{figure*}

\subsection{Preliminaries}

\subsubsection{Architecture of Image SR Baseline}

The architecture of quantized SR models is initiated by discussing its principal components. These models are designed to accept a low-resolution image, referred to as $I_{\text{LR}}$, and output a high-resolution counterpart, designated as $I_{\text{SR}}$. This transformation is mathematically encapsulated by:
\begin{equation}
I_{\text{SR}} = \mathcal{M}(I_{\text{LR}}),
\end{equation}
where $\mathcal{M}(\cdot)$ symbolizes the quantized SR model. The model $\mathcal{M}(\cdot)$ typically integrates three key elements: a low-level feature extractor $\mathcal{E}_{\text{L}}(\cdot)$, a high-level feature extractor $\mathcal{E}_{\text{H}}(\cdot)$, and a reconstruction module $\mathcal{R}(\cdot)$. The high-level feature extractor, which is the primary target for quantization~\cite{li2020pams}, is the most computationally intensive component of the network. The operational framework of the SR process, as previously defined, can be restructured as:
\begin{equation}
I_{\text{SR}} = \mathcal{R} \circ \mathcal{E}_{\text{H}} \circ \mathcal{E}_{\text{L}}(I_{\text{LR}}),
\end{equation}
where $\mathcal{E}_{\text{H}}$ denotes the high-level feature extractor, and $\circ$ indicates the sequential composition of model parts.

\subsubsection{Low-bit Quantization Framework}

In image SR models employing quantization, both weights $\boldsymbol{w}$ and activations $\boldsymbol{a}$ are condensed into reduced bit-width representations via respective quantizers, \textit{i.e.}, $Q_w(\boldsymbol{w})$ and $Q_a(\boldsymbol{a})$. This quantization commonly adopts a symmetric approach where the quantizers operate at a bit-width $b$, formulated as~\cite{wu2020integer}:
\begin{equation}
\label{eq:normal_quant}
Q^b_x(\boldsymbol x) = \left\lfloor\operatorname{clip}\left(\frac{\boldsymbol x}{v_x^b}, -1, 1\right)\right\rceil \times v_x^b,
\end{equation}
where $\boldsymbol x$ denotes the latent weight $\boldsymbol{w}$ or activation $\boldsymbol{a}$,
the clipping function $\operatorname{clip}(\cdot, -1, 1) = \max(\min(\cdot, 1), -1)$ restricts the input range.
Here, $v_x^b$ serves as the scaling factor, transforming higher precision values to their quantized lower bit-width equivalents, defined by $v^b_x = \frac{2\max\left(\left|\boldsymbol x\right|\right)}{2^b - 1}$.
The main computation of the quantized unit can be expressed as
\begin{equation}
\boldsymbol o = Q^b_w(\boldsymbol w) \otimes Q^b_a(\boldsymbol a) ,
\end{equation}
where $\boldsymbol o$ denotes the output and $\otimes$ denotes the convolution or linear projection composed of integer instructions.
Due to the non-differentiability of the quantizer, which would result in zero gradients and thereby hinder the backward propagation, the gradient approximation through straight-through estimation (STE)~\cite{bengio2013estimating} is utilized:
\begin{equation}
\frac{\partial Q^b(\boldsymbol x)}{\partial \boldsymbol x} =
\begin{cases}
1 & \text{if } \boldsymbol{x} \in (-r_x, r_x)\\
0 & \text{otherwise}
\end{cases}.
\end{equation}
Quantization in image SR models significantly reduces storage demands and computational burden by lowering bit widths and optimizing integer operations, thus enhancing operational efficiency.

\subsection{QuantSR+: Redistribution-driven Bit Determination}

\subsubsection{Quantization-induced Operator Degradation}
The quantization approach compresses weights and activations of models to low-bit representations for compact storage and accelerated computation, thus benefiting efficiency. However, quantization also results in a significant drop in accuracy in image SR tasks, especially when reducing model precision to ultra-low bit-widths (2-4 bits).
Specifically, the primary cause of this phenomenon is the high level of discretization introduced by the quantization function, which affects both the forward and backward propagation.

\textcolor{black}{During forward propagation, quantization makes feature representations much more discrete and less diverse, which can limit how precisely an SR network encodes fine-grained image details. For instance, moving from 32-bit floating-point to 2-bit quantization reduces the per-element value space from $2^{32}$ to $2^2$, creating an enormous representational gap of about $2^{30}$ (roughly one billion) times.
During backward propagation, quantizers are non-differentiable because they contain rounding and clipping operations. Therefore, training uses the straight-through estimator (STE)~\cite{bengio2013estimating}, which replaces the true derivative of the quantizer with an identity or clipped identity function. This surrogate gradient ignores quantization bin boundaries and assigns non-zero gradients even where the forward mapping is piecewise constant and has zero true derivative almost everywhere. As a result, the backward gradient is biased with respect to the true sensitivity of the quantized forward operator, creating a systematic forward-backward inconsistency. In low-bit SR settings, where quantization intervals are coarse, this inconsistency becomes more pronounced and can delay parameter updates from matching the true sensitivity changes that occur when activations or weights cross quantization bin boundaries, which directly affects reconstruction fidelity. Since STE is required for practical optimization of non-differentiable quantizers, this mismatch cannot be fully removed and must instead be mitigated by better quantization-aware training.}

\subsubsection{Redistribution-driven Bit Determination for Representation Recovery}
\begin{figure}
\centering
\includegraphics[width=\textwidth]{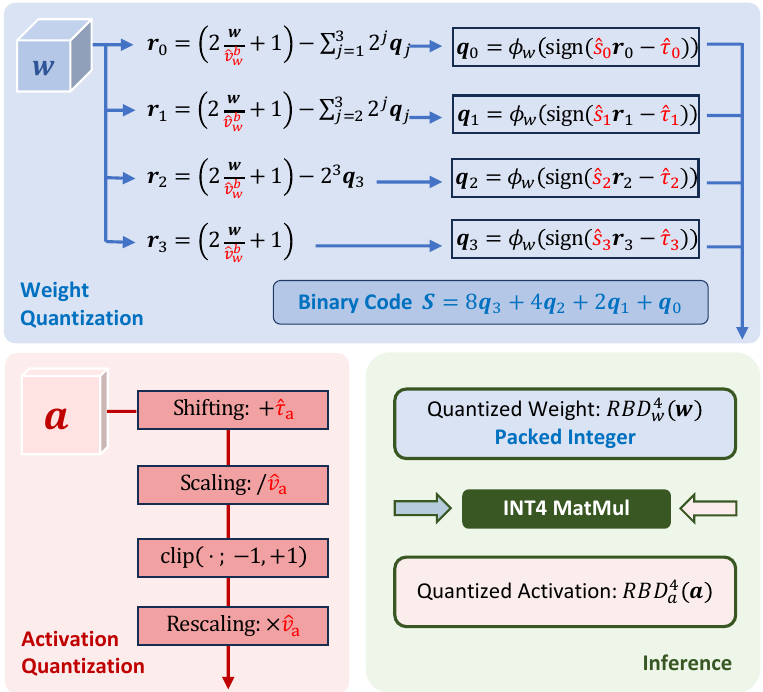}
\caption{\textcolor{black}{Training and inference processes of 4-bit RBD. \textcolor{black}{Red} notations are learnable parameters in RBD quantizers.}}
\label{fig:RBD}
\end{figure}
In this work, we introduce a novel Redistribution-driven Bit Determination (RBD) method for QuantSR+ (see Fig.~\ref{fig:RBD}).
\textcolor{black}{RBD stems from a straightforward motivation: to extremely fine-grain the forward and backward processes of quantization, building redistribution-driven quantizers at the single-bit level. During inference, RBD doesn't introduce any overhead and is fully consistent with standard hardware quantization operators. Ultimately, it facilitates accurate and flexible model optimization in SR tasks.}

\paragraph{RBD in Weight Quantization}

\textcolor{black}{In weight quantization, RBD introduces a learnable redistribution mechanism, \textit{i.e.}, a set of trainable parameters that control how much each binary bit contributes to the final quantized value. Specifically, it does \textbf{not} redistribute the original weight distribution directly; instead, it learns the relative contribution of different bit planes when composing an $b$-bit code. To this end, we reconstruct the weight quantizer by explicitly combining multiple binary bits into an $b$-bit integer-equivalent representation. Given a full-precision weight tensor $\boldsymbol{w}$, the $b$-bit quantized weight $\textit{RBD}^b_w(\boldsymbol{w})$ can be equivalently written as a weighted sum of $b$ binary bits with a shared channel-wise scaling factor. The formulation is}
\begin{equation}
\label{eq:rbd_w}
\textcolor{black}{
\textit{RBD}^b_w(\boldsymbol{w}) = \frac{1}{2} \left( S - 1 \right)\hat{v}_w,\quad
S = \sum_{i=0}^{b-1} 2^{i}\boldsymbol{q}_{i},}
\end{equation}
\textcolor{black}{where $\boldsymbol{q}_{i}\in\{-1,1\}$ denotes the $i$-th bit obtained by the sign function. $\hat{v}_w$ is a learnable channel-wise scaling factor that aligns the magnitude of the discrete codebook to that of the full-precision weights.}
\textcolor{black}{We initialize $\hat{v}_w$ by matching the mean absolute value between $\boldsymbol{w}$ and the integer code $S$, \textit{i.e.},}
\begin{equation}
\textcolor{black}{
\hat{v}_w \;\leftarrow\; \frac{2\,\mathbb{E}\!\left(|\boldsymbol{w}|\right)}{\mathbb{E}\!\left(|S-1|\right)},}
\end{equation}
\textcolor{black}{where $\mathbb{E}(\cdot)$ denotes the channel-wise average.}

Then, we design redistribution-driven determination functions for each bit.
\textcolor{black}{Specifically, we determine the $i$-th bit from a residual signal constructed in the integer-code domain. By removing the contributions of the already-determined higher bits, each decision focuses on the remaining quantization error. The proposed residual-driven bit determination can be written as:}
\begin{equation}
\label{eq:quantized_residual}
\textcolor{black}{
\boldsymbol{r}_i \;=\; \left( 2\frac{\boldsymbol{w}}{\hat{v}_w} + 1 \right)
\;-\; \sum\limits_{j=i+1}^{b-1} 2^{j}\boldsymbol{q}_{j},}
\end{equation}
\begin{equation}
\label{eq:quantized_bit}
\textcolor{black}{
\boldsymbol{q}_{i} \;=\; \operatorname{sign}\!\left( \phi_w\!\left( \hat{s}_i \boldsymbol{r}_i - \hat{\tau}_{w,i} \right)\right).}
\end{equation}
\textcolor{black}{where $\boldsymbol{r}_i$ denotes the residual before deciding $\boldsymbol{q}_i$ in the integer-code domain. It is obtained by first mapping the normalized weight to $\left(2\frac{\boldsymbol{w}}{\hat{v}_w}+1\right)$, which is consistent with the form in Eq.~\eqref{eq:rbd_w}. Then it subtracts the reconstructed contribution of higher bits $\{\boldsymbol{q}_{j}\}_{j>i}$.} $\hat{v}_w$, $\hat{s}$, and $\hat{\tau}_w$ are all learnable parameters. $\hat{v}_w$ adjusts the overall range of $\boldsymbol{w}$, while $\hat{s}$ and $\hat{\tau}_w$ regulate the distribution of the latent weights when determining each bit, initialized as 1 and 0, respectively. \textcolor{black}{By conditioning each bit on $\boldsymbol{r}_i$, the quantizer performs an explicit residual refinement: higher bits explain the dominant component first, and lower bits progressively compensate the remaining residual.} Simultaneously, the uniform application of the sign function yields a consistent post-discretization representation, so the resulting quantized outputs are equivalent to integer-quantized values and can be computed efficiently using standard integer instructions during inference.

We also introduce an embedded transformation function $\phi_w(\cdot)$ before the discretization in each bit determination function. The forward and backward processes are
\begin{equation}
\phi_w(\boldsymbol{x}) = \tanh(2\boldsymbol{x}),
\quad
{\partial\phi_w(\boldsymbol{x})}/{\partial \boldsymbol{x}} = 2\,\operatorname{sech}^2(2\boldsymbol{x}),
\end{equation}
\textcolor{black}{where $\boldsymbol{x}$ denotes the input to the external $\operatorname{sign}(\cdot)$ operator. Since $\tanh(\cdot)$ is strictly monotonic and odd, it preserves the sign of non-zero inputs, i.e., $\operatorname{sign}(\phi_w(\boldsymbol{x}))=\operatorname{sign}(\boldsymbol{x})$. In backward propagation, $2\,\operatorname{sech}^2(2\boldsymbol{x})$ smoothly down-weights gradients for elements with large $|\boldsymbol{x}|$ that are far from the decision boundary, while maintaining well-defined, non-zero gradients for finite inputs. This encourages stable optimization by focusing updates on the latent weights that are more likely to affect the discretized bit decisions, thereby better reflecting the forward quantization behavior.}

\paragraph{RBD in Activation Quantization}

For the quantization of activations, the ongoing variation between different propagation cycles during inference results in computationally expensive per-bit decisions. Therefore, we treat the quantizer for activations as an integrated integer quantizer, which is defined as follows:
\begin{equation}
\textit{RBD}^b_a(\boldsymbol{a}) = \left\lfloor\phi_a\left( \operatorname{clip}\left({\left(\boldsymbol{a} + \hat{\tau}_a\right)}/{\hat{v}_a} , -1, 1\right)\right)\right\rceil \times \hat{v}_a,
\end{equation}
where $\hat{v}_a$ and $\hat{\tau}_a$ denote learnable scaling factors and mean-shifts for quantized activations, respectively, which are both learnable parameters.
The formulation allows the activation quantizer to adaptively adjust the range and granularity of the quantization intervals, leveraging redistribution to preserve the diversity of the activations and ensure accurate representation even at lower bit widths:
\begin{equation}
\phi_a(\boldsymbol{a})={\tanh\left(2(\boldsymbol{a} - \lfloor\boldsymbol{a}\rfloor)-1\right)}/{\tanh{(1)}} + \lfloor\boldsymbol{a}\rfloor + 1,
\end{equation}
\begin{equation}
\label{eq:rbd_a}
{\partial \phi_a(\boldsymbol{a})}/{\partial \boldsymbol{a}} = {2\,\operatorname{sech}^2\bigl(2(\boldsymbol{a} - \lfloor \boldsymbol{a} \rfloor)-1\bigr)}/{\tanh(1)}.
\end{equation}
\textcolor{black}{This function does not alter the rounding value, but mitigates the gradient of elements distant from the center within each interval. It thus better reflects the quantization behavior and achieves more accurate yet stable gradients.}

\textcolor{black}{The proposed RBD quantizer improves quantized weights and activations in both forward and backward propagation. In the \textbf{forward pass}, bit decisions are modulated by learnable redistribution parameters, which adapt the contribution of each bit plane during training and yield accurate yet flexible low-bit representations. Notably, the resulting discrete representations are inherently integer-equivalent under the bit-composition form in Eq.~\eqref{eq:rbd_w}, and thus can be directly realized by standard integer quantization operators for deployment. In the \textbf{backward pass}, the embedded transformation functions $\phi_w(\cdot)$ and $\phi_a(\cdot)$ provide smooth, content-aware surrogate gradients. Compared with a plain STE, this design down-weights gradients for elements far from decision boundaries, leading to more stable and effective parameter updates without adding inference overhead. During \textbf{inference}, the integer-equivalent weights $\textit{RBD}^b_w(\boldsymbol{w})$ and quantized activations $\textit{RBD}^b_a(\boldsymbol{a})$ are multiplied using integer arithmetic for efficient execution.}

\subsection{QuantSR+: Quantized Slimmable Architecture}

\subsubsection{Constraints from Latent Structures}
\begin{figure}
\centering
\includegraphics[width=1.\textwidth]{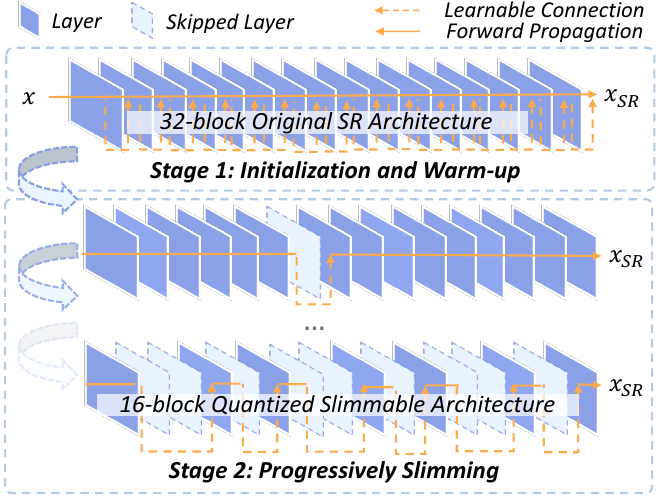}
\caption{The illustration of QSA. A dual-stage approach is applied to progressively slim and optimize the quantized SR model.}
\label{fig:DQA}
\end{figure}
As a model compression method utilizing low-bit parameters and computations, quantization enables efficient inference of SR models on resource-constrained hardware. Specifically, one of the main advantages of quantization is its universality regarding the bit-width of parameters in the model. This means it is adaptable to diverse architectures, thus enabling the compression of various SR models without altering their fundamental structures.

\textcolor{black}{However, the architectural universality of quantization also introduces two key limitations for quantized SR models. First, the performance ceiling is often bounded by the full-precision counterpart with the same fixed structure and parameter budget, inevitably leaving an accuracy gap. Second, quantization alone does not remove structural redundancy, implying that more compact structures may exist under the same accuracy target. Our conference version, QuantSR~\cite{qin2024quantsr}, has explored slimming, but it adopts a one-shot slimming at a specific point during training. This abrupt change in network topology can perturb optimization, and it may also invalidate part of the optimization progress accumulated before slimming, since the pre-slim parameters/representations cannot be consistently aligned with the post-slim architecture.}

\textcolor{black}{To address these issues, QuantSR+ introduces a Quantized Slimmable Architecture (QSA) that evolves progressively during training, rather than shrinking the model in a single step. By warm-starting from an over-parameterized quantized network with learnable shortcuts and then gradually skipping less important blocks, QSA preserves training continuity, reuses previously learned representations, and yields a more stable optimization path toward the final slim and efficient architecture.}

\subsubsection{Quantized Slimmable Architecture for Efficient Inference}
We introduce the Quantized Slimmable Architecture (QSA) for QuantSR+, specifically tailored for the advanced feature extractor $\mathcal{E}_\text{H}'$ in quantized SR models (see Figure ~\ref{fig:DQA}). Compared with the previous methods with direct quantization or one-shot slimming, the proposed QSA further pushes the limit of the representation capacity from a structure perspective to achieve higher accuracy.

Our first step is to construct the initialization of the slimmable architecture. The vast majority of mainstream architectures used for SR tasks, including convolution-based~\cite{ledig2017photo} and transformer-based~\cite{liang2021swinir} models, consist of several computational blocks of the same size. Considering their quantization-aware training process, the latent weights in different blocks of the model exhibit varying sensitivities to quantization~\cite{Dong2019b,jhunjhunwala2021adaquant}, and extracted features have different levels of importance for prediction. Therefore, we first construct an initial setting for the quantized SR model with twice the number of blocks, allowing it to surpass the original scale's accuracy upper limit after quantization. This structure is based on quantized blocks accompanied by learnable shortcut connections, represented as $\mathcal{E}_\text{H}^\text{QSA}$, composed of $2N$ blocks ($N \in \{2^n, n\in\mathbf{Z}^+\}$), where the initial configuration of the $i$-th block $\Phi_i^\text{QSA}(\cdot)$ is described as follows:
\begin{equation}
\Phi_i^\text{QSA} (\boldsymbol{x}_i) = \varphi_i (\boldsymbol{x}_i) + \alpha_i\boldsymbol{x}_i,
\end{equation}
where $\boldsymbol{x}_i$ represents the input features of the $i$-th block, $\varphi(\cdot)$ denotes the quantized feature extractor consisting of convolutional and activation layers, and $\alpha$ is the learnable scaling factor of the shortcut connections.
We perform a warm-up process for the constructed initial architecture, which only accounts for 20\% of the total number of iterations in the training process. During the warm-up phase, $\varphi_i (\cdot)$ within $\Phi_i^\text{QSA}(\cdot)$ undergoes task-oriented optimization, while $\alpha_i$ is also updated during the learning process. Throughout the model, a lower $\alpha_i$ implies that the corresponding block's computation is more difficult to disregard, signifying its greater importance within the entire quantized architecture.

\textcolor{black}{Following initialization, QSA tightly integrates slimming with quantization-aware training and evolves the architecture in a progressive manner, rather than switching the network to a final slim topology at a specific stage~\cite{qin2024quantsr}. Specifically, after the warm-up phase, we iteratively skip the blocks with the highest $\alpha$ (remove them from the set $S$) and continue training the resulting student until it recovers the pre-slim accuracy. The skipped blocks of QSA will be removed in deployed inference. This procedure is repeated until the target efficiency constraint is satisfied (e.g., retaining $N$ blocks, i.e., skipping 50\% blocks), yielding a final architecture that is reached smoothly through successive intermediate states. The final QSA structure can be represented as:}
\begin{equation}
\small
\mathcal{E}_\text{H}^\text{QSA}(x, S^*) = \prod_{i=1}^{2N} \left[\Phi_i^\text{QSA} \in S^*\right]
\Phi_i^\text{QSA} (\boldsymbol{x}_i) + \left[\Phi_i^\text{QSA} \not\in S^*\right] \boldsymbol{x}_i,
\end{equation}
where $[\cdot]$ denotes the Iverson bracket, equaling 1 if the condition inside the bracket is true, $S^*$ represents the set of blocks not skipped in the final slim architecture, and $\prod$ denotes the sequential combination of quantized blocks within the network. After the evolution of the structure, we train the resulting QSA until the complete quantization-aware training process concludes.
In practice, we cap the number of iterations for each intermediate evolutionary state to avoid stalling at a state that fails to satisfy the target constraints. Specifically, the iteration budget for each state is set to $\frac{1}{N} \times 80\%$ of the total number of iterations.

\textcolor{black}{QSA makes QuantSR+ both accurate and efficient by coupling quantization-aware training with a continuous architectural evolution: (1) In initialization, an over-parameterized (2$N$-block) quantized backbone with learnable shortcuts improves representation capacity and provides an explicit importance signal for each block during warm-up. (2) In slimming, QSA progressively skips the least critical blocks according to these learned importance factors, yielding a compact $N$-block network while preserving training continuity. Compared with our conference method QuantSR~\cite{qin2024quantsr}, which performs one-shot slimming at a fixed training stage, QSA avoids abrupt topology changes that perturb optimization and invalidate previously learned representations, leading to a more stable path to the final efficient model.}

\subsection{QuantSR+: Slimming-guided Function-localized Distillation}

\subsubsection{Optimization Perturbations in Distillation}
\begin{figure}
\centering
\includegraphics[width=1.\textwidth]{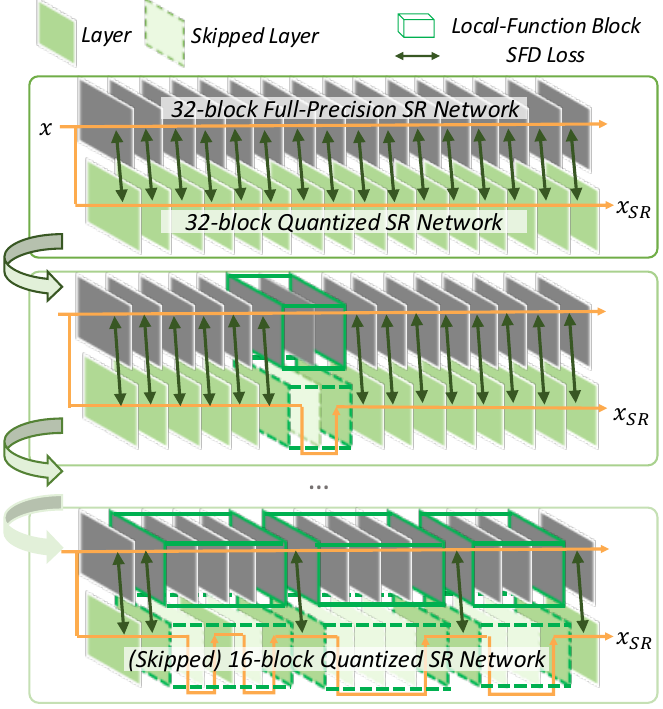}
\caption{SFD integrates different layers to form local functional blocks corresponding to full-precision and quantized SR models to accelerate the convergence of quantized SR models.}
\label{fig:SFD}
\end{figure}
The high discreteness introduced by quantization interferes with the optimization of gradient-based SR models, not only due to the limited representation caused by compression in the forward propagation but also due to the inevitable errors brought by the gradient approximation.
Due to the availability of a full-precision counterpart, distillation is considered an intuitive method to improve the performance of quantized models and has been widely used to optimize SR models. Specifically, quantization-aware distillation compares the outputs of a quantized SR student model and a full-precision teacher SR model to optimize the performance of the former. Moreover, when the student network is the full-precision counterpart of the teacher network, distillation can be performed at a finer granularity, such as layer-wise or block-wise (multi-layer) distillation.

However, the proposed QSA has an architecture that dynamically evolves during the optimization process, making it challenging to utilize existing quantization-aware distillation frameworks for SR models. After architectural changes, distillation can only be performed at the output positions of the SR model, or the full-precision counterpart of the evolved architecture needs to be retrained. This means that the training and optimization efficiency under these frameworks is severely limited. Therefore, we propose that QuantSR+ with QSA requires a new quantization-aware distillation method to enable universal and efficient optimization for quantized models in image SR tasks.

\subsubsection{Slimming-guided Function-localized Distillation for Stable Convergence}
We introduce the Slimming-guided Function-localized Distillation (SFD) to optimize dynamic quantized SR networks, which is shown in Fig.~\ref{fig:SFD}. SFD allows for the optimization process to exploit a powerful, full-size, full-precision SR teacher to supervise a slim, ultra-low-bit quantized student model, utilizing a function-localized strategy to accelerate the convergence of the quantized model.

Throughout the training process, the Quantized Slimming Architecture (QSA) evolves from an initial full-size (2N blocks) structure to a slim (N blocks) final version. We first propose a foundational framework for the distillation of the quantized architecture, specifically the block-to-block correspondence distillation between the full-size teacher and student. Given the same representation dimensions and number of blocks, we directly compute the Mean Squared Error (MSE) loss at the block level to optimize the model:
\begin{equation}
\label{eq:l_sfd}
\mathcal{L}_{\text{distill}}^{\text{init}} = \frac{1}{2N} \sum_{i=1}^{2N} \left\| \frac{\Phi_i^{\text{FP}} (\boldsymbol{x}_i^{\text{FP}}) - \Phi_i^{\text{QSA}} (\boldsymbol{x}_i)}{\left|\Phi_i^{\text{FP}} (\boldsymbol{x}_i^{\text{FP}})\right|} \right\|_{\ell2}^2,
\end{equation}
where $\Phi_i^{\text{FP}}$ and $\Phi_i^{\text{QSA}}$ denote the output representation of the $i$-th block from full-precision and quantized SR models, respectively, $\|\cdot\|_{\ell2}$ denotes the $\ell2$ normalization function, and $|\cdot|$ denotes the cardinality function of the set, \textit{i.e.}, the number of elements in the set.

After the onset of evolution, we propose a slimming-guided, function-localized distillation strategy between the full-size teacher and the slim student. The aim of this strategy is to establish a simple yet effective correspondence to leverage the quantized blocks surrounding the slimmed blocks to compensate for the local functionality of the slimmed blocks, allowing most of the blocks in the quantized SR model to be under stable supervision and optimization during evolution.
Consider the first slimming of the initial quantized SR student. After skipping the $i$-th block, the supervision of the input (if any) and output of the $i$-th block is removed from the loss, retaining the distillation loss items established by the input of the $\{i-1\}$-th block and the output of the $\{i+1\}$-th block with the teacher model. This means that we expect the set of blocks {$\Phi_{i-1}^\text{QSA}$, $\Phi_{i+1}^\text{QSA}$} in the quantized model to learn the corresponding functionality of the block set {$\Phi_{i-1}^\text{FP}$, $\Phi_{i}^\text{FP}$, $\Phi_{i+1}^\text{FP}$} in the full-precision model, thereby producing similar outputs at $\{i+1\}$-th block.
Supervision of other blocks in the quantized model remains unchanged, allowing them to learn stably throughout the optimization process as long as they are not slimmed. The proposed SFD can be represented as follows:
\begin{equation}
\begin{aligned}
\mathcal{L}_\text{distill}^\text{SFD}=\frac{1}{2N}\sum_{i=1}^{2N}&\left[\Phi_i^\text{QSA} \in S\right]\cdot\left[\Phi_{i+1}^\text{QSA} \in S\right]\cdot\\
&\left\|\frac{\Phi_i^\text{FP} (\boldsymbol{x}^\text{FP}_i)-\Phi_i^\text{QSA} (\boldsymbol{x}_i)}{\left|\Phi_i^\text{FP} (\boldsymbol{x}^\text{FP}_i)\right|}\right\|_2^2.
\end{aligned}
\end{equation}
In the SFD for training optimization, the set of skipped blocks $S$ dynamically changes (until the final optimized $S^*$), meaning that supervision is modified as the architecture evolves. However, for all non-adjacent skipped blocks currently within the set $S$, the supervision at their input and output positions does not change during optimization. This allows them to undergo stable optimization towards a consistent objective, enabling faster achievement of a highly accurate quantized SR model.

\begin{table*}[!h]
\small
\centering
\resizebox{1\columnwidth}{!}{
\begin{tabular}{lcccccccccccc}
\hline
\rowcolor{colorTabTop}
Method & Scale & \#Bit & \multicolumn{2}{c}{Set5} & \multicolumn{2}{c}{Set14} & \multicolumn{2}{c}{B100} & \multicolumn{2}{c}{Urban100} & \multicolumn{2}{c}{Manga109} \\
\cline{4-13}
\rowcolor{colorTabTop}
& & ($w/a$) & PSNR & SSIM & PSNR & SSIM & PSNR & SSIM & PSNR & SSIM & PSNR & SSIM \\
\hline
\hline
Bicubic & $\times$2 & -/- & 33.66 & 0.9299 & 30.24 & 0.8688 & 29.56 & 0.8431 & 26.88 & 0.8403 & 30.80 & 0.9339 \\
SRResNet~\cite{ledig2017photo} & $\times$2 & 32/32 & 38.00 & 0.9605 & 33.59 & 0.9171 & 32.19 & 0.8997 & 32.11 & 0.9282 & 38.56 & 0.9770 \\
DoReFa~\cite{zhou2016dorefa} & $\times$2 & 8/8 & 37.32 & 0.9520 & 32.90 & 0.8680 & 31.69 & 0.8504 & 30.32 & 0.8800 & 37.01 & 0.9450 \\
CADyQ~\cite{hong2022cadyq} & $\times$2 & 8/8 & 37.79 & 0.9590 & 33.37 & 0.9150 & 32.02 & 0.8980 & 31.53 & 0.9230 & 38.06 & 0.9760 \\
\textcolor{black}{AdaBM~\cite{hong2024adabm}} & \textcolor{black}{$\times$2} & \textcolor{black}{4/4MP} & \textcolor{black}{37.10} & \textcolor{black}{0.9550} & \textcolor{black}{32.85} & \textcolor{black}{0.9100} & \textcolor{black}{31.63} & \textcolor{black}{0.8910} & \textcolor{black}{30.48} & \textcolor{black}{0.9120} & \textcolor{black}{-} & \textcolor{black}{-} \\
\textcolor{black}{CABM~\cite{tian2023cabm}} & \textcolor{black}{$\times$2} & \textcolor{black}{4/4MP} & \textcolor{black}{-} & \textcolor{black}{-} & \textcolor{black}{-} & \textcolor{black}{-} & \textcolor{black}{-} & \textcolor{black}{-} & \textcolor{black}{31.54} & \textcolor{black}{0.9230} & \textcolor{black}{-} & \textcolor{black}{-} \\
\textcolor{black}{Granular-DQ~\cite{wang2025thinking}} & \textcolor{black}{$\times$2} & \textcolor{black}{4/4MP} & \textcolor{black}{-} & \textcolor{black}{-} & \textcolor{black}{-} & \textcolor{black}{-} & \textcolor{black}{-} & \textcolor{black}{-} & \textcolor{black}{31.94} & \textcolor{black}{0.9270} & \textcolor{black}{-} & \textcolor{black}{-} \\
\hline
DoReFa~\cite{zhou2016dorefa} & $\times$2 & 4/4 & 37.31 & 0.9510 & 32.48 & 0.9091 & 31.64 & 0.8901 & 30.18 & 0.8780 & 36.95 & 0.9440 \\
PAMS~\cite{li2020pams} & $\times$2 & 4/4 & 37.67 & 0.9588 & 33.19 & 0.9146 & 31.90 & 0.8966 & 31.10 & 0.9194 & 37.62 & 0.9740 \\
CADyQ~\cite{hong2022cadyq} & $\times$2 & 4/4 & 37.58 & 0.9580 & 33.14 & 0.9140 & 31.87 & 0.8960 & 30.94 & 0.9170 & 37.31 & 0.9740 \\
QuantSR~\cite{qin2024quantsr} & $\times$2 & 4/4 & 37.80 & 0.9597 & 33.35 & 0.9158 & 32.04 & 0.8979 & 31.46 & 0.9221 & 38.25 & 0.9762 \\
\textcolor{black}{PTQ4SR~\cite{tu2023toward}} & \textcolor{black}{$\times$2} & \textcolor{black}{4/4} & \textcolor{black}{36.49} & \textcolor{black}{0.9510} & \textcolor{black}{32.40} & \textcolor{black}{0.9040} & \textcolor{black}{31.36} & \textcolor{black}{0.8850} & \textcolor{black}{29.90} & \textcolor{black}{0.9040} & \textcolor{black}{-} & \textcolor{black}{-} \\
\textcolor{black}{I$^2$NQ~\cite{sun20242} $^*$} & \textcolor{black}{$\times$2} & \textcolor{black}{4/4} & \textcolor{black}{37.94} & \textcolor{black}{0.9603} & \textcolor{black}{33.54} & \textcolor{black}{0.9168} & \textcolor{black}{32.10} & \textcolor{black}{0.8988} & \textcolor{black}{31.74} & \textcolor{black}{0.9250} & \textcolor{black}{38.48} & \textcolor{black}{0.9768} \\
\rowcolor{colorTabTop}
QuantSR+ & $\times$2 & 4/4 & {37.91} & {0.9602} & {33.52} & {0.9165} & {32.09} & {0.8986} & {31.75} & {0.9250} & {38.46} & {0.9767} \\
\rowcolor{colorTabTop}
\textcolor{black}{QuantSR+$^*$} & \textcolor{black}{$\times$2} & \textcolor{black}{4/4} & \textcolor{black}{37.96} & \textcolor{black}{0.9605} & \textcolor{black}{33.60} & \textcolor{black}{0.9172} & \textcolor{black}{32.15} & \textcolor{black}{0.8990} & \textcolor{black}{31.79} & \textcolor{black}{0.9254} & \textcolor{black}{38.55} & \textcolor{black}{0.9770} \\
\hline
DoReFa~\cite{zhou2016dorefa} & $\times$2 & 2/2 & 36.91 & 0.9470 & 32.55 & 0.9071 & 31.41 & 0.8868 & 29.60 & 0.8740 & 36.13 & 0.9410 \\
PAMS~\cite{li2020pams} & $\times$2 & 2/2 & 34.04 & 0.8270 & 30.91 & 0.8751 & 30.11 & 0.8592 & 27.57 & 0.8400 & 31.79 & 0.9110 \\
CADyQ~\cite{hong2022cadyq} & $\times$2 & 2/2 & 19.44 & 0.5610 & 18.51 & 0.4810 & 19.70 & 0.4760 & 17.97 & 0.4550 & 17.35 & 0.5830 \\
QuantSR~\cite{qin2024quantsr} & $\times$2 & 2/2 & 37.57 & 0.9589 & 33.09 & 0.9136 & 31.84 & 0.8954 & 30.77 & 0.9149 & 37.60 & 0.9745 \\
\rowcolor{colorTabTop}
QuantSR+ & $\times$2 & 2/2 & {37.77} & {0.9596} & {33.32} & {0.9147} & {31.97} & {0.8968} & {31.25} & {0.9198} & {37.96} & {0.9756} \\
\hline
\hline
Bicubic & $\times$4 & -/- & 28.42 & 0.8104 & 26.00 & 0.7027 & 25.96 & 0.6675 & 23.14 & 0.6577 & 24.89 & 0.7866 \\
SRResNet~\cite{ledig2017photo} & $\times$4 & 32/32 & 32.16 & 0.8951 & 28.60 & 0.7822 & 27.58 & 0.7364 & 26.11 & 0.7870 & 30.46 & 0.9089 \\
\textcolor{black}{AdaBM~\cite{hong2024adabm}} & \textcolor{black}{$\times$4} & \textcolor{black}{4/4MP} & \textcolor{black}{31.02} & \textcolor{black}{0.8600} & \textcolor{black}{27.87} & \textcolor{black}{0.7510} & \textcolor{black}{26.91} & \textcolor{black}{0.7000} & \textcolor{black}{25.11} & \textcolor{black}{0.7360} & \textcolor{black}{-} & \textcolor{black}{-} \\
\textcolor{black}{CABM~\cite{tian2023cabm}} & \textcolor{black}{$\times$4} & \textcolor{black}{4/4MP} & \textcolor{black}{-} & \textcolor{black}{-} & \textcolor{black}{-} & \textcolor{black}{-} & \textcolor{black}{-} & \textcolor{black}{-} & \textcolor{black}{25.86} & \textcolor{black}{0.7780} & \textcolor{black}{-} & \textcolor{black}{-} \\
\textcolor{black}{RefQSR ($\delta$-4bit)~\cite{lee2024refqsr}} & \textcolor{black}{$\times$4} & \textcolor{black}{4/4MP} & \textcolor{black}{-} & \textcolor{black}{-} & \textcolor{black}{-} & \textcolor{black}{-} & \textcolor{black}{-} & \textcolor{black}{-} & \textcolor{black}{25.90} & \textcolor{black}{0.7780} & \textcolor{black}{-} & \textcolor{black}{-} \\
\textcolor{black}{Granular-DQ~\cite{wang2025thinking}} & \textcolor{black}{$\times$4} & \textcolor{black}{4/4MP} & \textcolor{black}{-} & \textcolor{black}{-} & \textcolor{black}{-} & \textcolor{black}{-} & \textcolor{black}{-} & \textcolor{black}{-} & \textcolor{black}{25.98} & \textcolor{black}{0.7830} & \textcolor{black}{-} & \textcolor{black}{-} \\
\hline
DoReFa~\cite{zhou2016dorefa} & $\times$4 & 4/4 & 29.57 & 0.8369 & 26.82 & 0.7352 & 26.47 & 0.6971 & 23.75 & 0.6898 & 27.89 & 0.8634 \\
PAMS~\cite{li2020pams} & $\times$4 & 4/4 & 31.59 & 0.8851 & 28.20 & 0.7725 & 27.32 & 0.7220 & 25.32 & 0.7624 & 28.86 & 0.8805 \\
CADyQ~\cite{hong2022cadyq} & $\times$4 & 4/4 & 31.48 & 0.8830 & 28.05 & 0.7690 & 27.21 & 0.7240 & 25.09 & 0.7520 & 28.82 & 0.8840 \\
\textcolor{black}{PTQ4SR~\cite{tu2023toward}} & \textcolor{black}{$\times$4} & \textcolor{black}{4/4} & \textcolor{black}{31.15} & \textcolor{black}{0.8780} & \textcolor{black}{27.89} & \textcolor{black}{0.7630} & \textcolor{black}{27.15} & \textcolor{black}{0.7180} & \textcolor{black}{25.13} & \textcolor{black}{0.7530} & \textcolor{black}{-} & \textcolor{black}{-} \\
QuantSR~\cite{qin2024quantsr} & $\times$4 & 4/4 & 32.00 & 0.8924 & 28.50 & 0.7799 & 27.52 & 0.7342 & 25.88 & 0.7807 & 30.15 & 0.9040 \\
\textcolor{black}{I$^2$NQ~\cite{sun20242} $^*$} & \textcolor{black}{$\times$4} & \textcolor{black}{4/4} & \textcolor{black}{32.16} & \textcolor{black}{0.8948} & \textcolor{black}{28.53} & \textcolor{black}{0.7809} & \textcolor{black}{27.56} & \textcolor{black}{0.7357} & \textcolor{black}{25.95} & \textcolor{black}{0.7826} & \textcolor{black}{30.30} & \textcolor{black}{0.9062} \\
\rowcolor{colorTabTop}
QuantSR+ & $\times$4 & 4/4 & {32.13} & {0.8940} & {28.50} & {0.7804} & {27.53} & {0.7351} & {25.94} & {0.7825} & {30.30} & {0.9061} \\
\rowcolor{colorTabTop}
\textcolor{black}{QuantSR+$^*$} & \textcolor{black}{$\times$4} & \textcolor{black}{4/4} & \textcolor{black}{32.17} & \textcolor{black}{0.8951} & \textcolor{black}{28.53} & \textcolor{black}{0.7810} & \textcolor{black}{27.60} & \textcolor{black}{0.7363} & \textcolor{black}{26.01} & \textcolor{black}{0.7832} & \textcolor{black}{30.35} & \textcolor{black}{0.9076} \\
\hline
DoReFa~\cite{zhou2016dorefa} & $\times$4 & 2/2 & 30.54 & 0.8610 & 27.50 & 0.7538 & 26.90 & 0.7098 & 24.44 & 0.7242 & 27.31 & 0.8502 \\
PAMS~\cite{li2020pams} & $\times$4 & 2/2 & 29.20 & 0.8239 & 26.61 & 0.7273 & 26.36 & 0.6934 & 23.58 & 0.6812 & 25.59 & 0.8012 \\
CADyQ~\cite{hong2022cadyq} & $\times$4 & 2/2 & 19.67 & 0.5380 & 19.30 & 0.4740 & 19.80 & 0.4620 & 17.97 & 0.4360 & 17.30 & 0.5640 \\
\textcolor{black}{DAQ~\cite{hong2022daq}} & \textcolor{black}{$\times$4} & \textcolor{black}{2/2} & \textcolor{black}{31.67} & \textcolor{black}{-} & \textcolor{black}{28.26} & \textcolor{black}{-} & \textcolor{black}{27.32} & \textcolor{black}{-} & \textcolor{black}{25.39} & \textcolor{black}{-} & \textcolor{black}{-} & \textcolor{black}{-} \\
QuantSR~\cite{qin2024quantsr} & $\times$4 & 2/2 & 31.30 & 0.8819 & 28.08 & 0.7694 & 27.23 & 0.7246 & 25.13 & 0.7537 & 28.81 & 0.8844 \\
\textcolor{black}{ODM~\cite{hong2024overcoming}} & \textcolor{black}{$\times$4} & \textcolor{black}{2/2} & \textcolor{black}{31.81} & \textcolor{black}{0.8880} & \textcolor{black}{28.32} & \textcolor{black}{0.7740} & \textcolor{black}{27.38} & \textcolor{black}{0.7300} & \textcolor{black}{25.54} & \textcolor{black}{0.7670} & \textcolor{black}{-} & \textcolor{black}{-} \\
\rowcolor{colorTabTop}
QuantSR+ & $\times$4 & 2/2 & {31.89} & {0.8908} & {28.37} & {0.7763} & {27.43} & {0.7308} & {25.58} & {0.7702} & {29.67} & {0.8974} \\
\hline
\end{tabular}
}
\caption{Quantitative results. SRResNet is used as the full-precision backbone. `$w/a$' denotes the weight/activation bits, and `MP' denotes the mixed-precision settings. $*$ denotes the activation shifting and partition in \cite{sun20242}.}
\label{tab:quantsr_srresnet}
\end{table*}

Considering the proposed SFD training strategy, we provide the overall training loss for QuantSR+. Given a training dataset $D = \left\{\mathcal I_\text{LR}^i, \mathcal I_\text{HR}^i\right\}_{i=1}^K$ consisting of $K$ low-resolution inputs $\mathcal I_\text{LR}$ and their corresponding high-resolution ground truths $\mathcal I_\text{HR}$, the traditional pixel loss $\mathcal{L}^\text{PIX}$ for the quantized SR model $\mathcal{M}$ can be expressed as:
\begin{equation}
\label{eq:joint_training}
\mathcal{L}^\text{PIX} = \frac{1}{K} \sum_{i=1}^K \left\| \mathcal I_\text{HR}^i - \mathcal{M}(\mathcal I_\text{LR}^i) \right\|_{\ell_1}.
\end{equation}
This pixel loss is jointly optimized with the SFD loss to train the network as follows:
\begin{equation}
\label{eq:total_loss}
\mathcal{L} = \mathcal{L}^\text{PIX} + \lambda \mathcal{L}^\text{SFD},
\end{equation}
where $\lambda$ is a balancing coefficient between the pixel loss and the SFD loss, empirically set to $1 \times 10^{-4}$.

With carefully designed quantizers, architectures, and optimization pipelines, QuantSR+ pushes the limits of accuracy and efficiency for quantized SR networks. Moreover, it can be universally applied to and improve backbones based on both convolution and attention mechanisms.

\begin{table*}[!h]
\small
\centering
\resizebox{1\columnwidth}{!}{
\begin{tabular}{lcccccccccccc}
\hline
\rowcolor{colorTabTop}
Method & Scale & \#Bit & \multicolumn{2}{c}{Set5} & \multicolumn{2}{c}{Set14} & \multicolumn{2}{c}{B100} & \multicolumn{2}{c}{Urban100} & \multicolumn{2}{c}{Manga109} \\
\cline{4-13}
\rowcolor{colorTabTop}
& & ($w/a$) & PSNR & SSIM & PSNR & SSIM & PSNR & SSIM & PSNR & SSIM & PSNR & SSIM \\
\hline
\hline
Bicubic & $\times$2 & -/- & 33.66 & 0.9299 & 30.24 & 0.8688 & 29.56 & 0.8431 & 26.88 & 0.8403 & 30.80 & 0.9339 \\
SwinIR\_S~\cite{liang2021swinir} & $\times$2 & 32/32 & 38.14 & 0.9611 & 33.86 & 0.9206 & 32.31 & 0.9012 & 32.76 & 0.9340 & 39.12 & 0.9783 \\
\hline
2DQuant~\cite{liu20252dquant} & $\times$2 & 4/4 & 37.87 & 0.9594 & 33.41 & 0.9161 & 32.02 & 0.8971 & 31.84 & 0.9251 & 38.31 & 0.9761 \\
QuantSR~\cite{qin2024quantsr} & $\times$2 & 4/4 & 38.10 & 0.9604 & 33.65 & 0.9186 & 32.21 & 0.8998 & 32.20 & 0.9295 & 38.85 & 0.9774 \\
\rowcolor{colorTabTop}
QuantSR+ & $\times$2 & 4/4 & 38.08 & 0.9608 & 33.86 & 0.9207 & 32.20 & 0.8999 & 32.31 & 0.9302 & 38.89 & 0.9776 \\
\hline
2DQuant~\cite{liu20252dquant} & $\times$2 & 2/2 & 36.00 & 0.9497 & 31.98 & 0.9012 & 30.91 & 0.8810 & 28.62 & 0.8819 & 34.40 & 0.9602 \\
QuantSR~\cite{qin2024quantsr} & $\times$2 & 2/2 & 37.55 & 0.9587 & 33.12 & 0.9143 & 31.89 & 0.8958 & 30.96 & 0.9172 & 37.61 & 0.9745 \\
\rowcolor{colorTabTop}
QuantSR+ & $\times$2 & 2/2 & 37.67 & 0.9591 & 33.25 & 0.9145 & 31.91 & 0.8960 & 31.06 & 0.9178 & 37.85 & 0.9750 \\
\hline
\hline
Bicubic & $\times$4 & -/- & 28.42 & 0.8104 & 26.00 & 0.7027 & 25.96 & 0.6675 & 23.14 & 0.6577 & 24.89 & 0.7866 \\
SwinIR\_S~\cite{liang2021swinir} & $\times$4 & 32/32 & 32.44 & 0.8976 & 28.77 & 0.7858 & 27.69 & 0.7406 & 26.47 & 0.7980 & 30.92 & 0.9151 \\
\hline
2DQuant~\cite{liu20252dquant} & $\times$4 & 4/4 & 31.77 & 0.8867 & 28.30 & 0.7733 & 27.37 & 0.7278 & 25.71 & 0.7712 & 29.71 & 0.8972 \\
\textcolor{black}{ODM~\cite{hong2024overcoming}} & \textcolor{black}{$\times$4} & \textcolor{black}{4/4} & \textcolor{black}{32.17} & \textcolor{black}{0.8920} & \textcolor{black}{28.59} & \textcolor{black}{0.7810} & \textcolor{black}{27.56} & \textcolor{black}{0.7360} & \textcolor{black}{26.06} & \textcolor{black}{0.7850} & \textcolor{black}{-} & \textcolor{black}{-} \\
QuantSR~\cite{qin2024quantsr} & $\times$4 & 4/4 & 32.18 & 0.8941 & 28.63 & 0.7822 & 27.59 & 0.7367 & 26.11 & 0.7871 & 30.49 & 0.9087 \\
\rowcolor{colorTabTop}
QuantSR+ & $\times$4 & 4/4 & 32.25 & 0.8944 & 28.65 & 0.7830 & 27.60 & 0.7376 & 26.18 & 0.7892 & 30.58 & 0.9096 \\
\hline
2DQuant~\cite{liu20252dquant} & $\times$4 & 2/2 & 29.53 & 0.8372 & 26.86 & 0.7322 & 26.46 & 0.6927 & 23.84 & 0.6912 & 26.07 & 0.8163 \\
QuantSR~\cite{qin2024quantsr} & $\times$4 & 2/2 & 31.53 & 0.8845 & 28.16 & 0.7715 & 27.28 & 0.7274 & 25.26 & 0.7609 & 29.06 & 0.8898 \\
\textcolor{black}{ODM~\cite{hong2024overcoming}} & \textcolor{black}{$\times$4} & \textcolor{black}{2/2} & \textcolor{black}{31.67} & \textcolor{black}{0.8850} & \textcolor{black}{28.23} & \textcolor{black}{0.7720} & \textcolor{black}{27.33} & \textcolor{black}{0.7280} & \textcolor{black}{25.36} & \textcolor{black}{0.7620} & \textcolor{black}{-} & \textcolor{black}{-} \\
\rowcolor{colorTabTop}
\textcolor{black}{QuantSR+} & \textcolor{black}{$\times$4} & \textcolor{black}{2/2} & \textcolor{black}{31.70} & \textcolor{black}{0.8860} & \textcolor{black}{28.28} & \textcolor{black}{0.7731} & \textcolor{black}{27.37} & \textcolor{black}{0.7277} & \textcolor{black}{25.65} & \textcolor{black}{0.7695} & \textcolor{black}{29.69} & \textcolor{black}{0.8968} \\
\hline
\end{tabular}
}
\caption{Quantitative results. SwinIR-S is used as the full-precision backbone.}
\label{tab:quantsr_swinir}
\end{table*}

\begin{table*}[!h]
\small
\centering
\resizebox{1\columnwidth}{!}{
\begin{tabular}{lccccccccccc}
\hline
\rowcolor{colorTabTop}
\textcolor{black}{Method} & \textcolor{black}{Scale} & \textcolor{black}{\#Bit ($w/a$)} & \textcolor{black}{PSNR$\uparrow$} & \textcolor{black}{SSIM$\uparrow$} & \textcolor{black}{LPIPS$\downarrow$} & \textcolor{black}{DISTS$\downarrow$} & \textcolor{black}{FID$\downarrow$} & \textcolor{black}{NIQE$\downarrow$} & \textcolor{black}{MUSIQ$\uparrow$} & \textcolor{black}{MANIQA$\uparrow$} & \textcolor{black}{CLIPIQA$\uparrow$} \\
\hline
\hline
\textcolor{black}{StableSR~\cite{wang2024exploiting}} & \textcolor{black}{$\times$4} & \textcolor{black}{32/32} & \textcolor{black}{23.28} & \textcolor{black}{0.5733} & \textcolor{black}{0.3115} & \textcolor{black}{0.2049} & \textcolor{black}{24.55} & \textcolor{black}{4.7825} & \textcolor{black}{65.70} & \textcolor{black}{0.6171} & \textcolor{black}{0.6767} \\
\hline
\textcolor{black}{PAMS~\cite{li2020pams}} & \textcolor{black}{$\times$4} & \textcolor{black}{4/4} & \textcolor{black}{23.11} & \textcolor{black}{0.5379} & \textcolor{black}{0.4225} & \textcolor{black}{0.2886} & \textcolor{black}{89.99} & \textcolor{black}{3.7643} & \textcolor{black}{60.02} & \textcolor{black}{0.5083} & \textcolor{black}{0.7213} \\
\textcolor{black}{QuantSR~\cite{qin2024quantsr}} & \textcolor{black}{$\times$4} & \textcolor{black}{4/4} & \textcolor{black}{23.37} & \textcolor{black}{0.5464} & \textcolor{black}{0.4203} & \textcolor{black}{0.2872} & \textcolor{black}{88.24} & \textcolor{black}{3.9060} & \textcolor{black}{58.27} & \textcolor{black}{0.4972} & \textcolor{black}{0.7176} \\
\rowcolor{colorTabTop}
\textcolor{black}{QuantSR+} & \textcolor{black}{$\times$4} & \textcolor{black}{4/4} & \textcolor{black}{23.43} & \textcolor{black}{0.5612} & \textcolor{black}{0.3965} & \textcolor{black}{0.2754} & \textcolor{black}{78.62} & \textcolor{black}{3.9905} & \textcolor{black}{57.52} & \textcolor{black}{0.4911} & \textcolor{black}{0.6942} \\
\hline
\textcolor{black}{PAMS~\cite{li2020pams}} & \textcolor{black}{$\times$4} & \textcolor{black}{2/2} & \textcolor{black}{17.88} & \textcolor{black}{0.3080} & \textcolor{black}{0.8730} & \textcolor{black}{0.4106} & \textcolor{black}{198.69} & \textcolor{black}{8.6302} & \textcolor{black}{39.66} & \textcolor{black}{0.3552} & \textcolor{black}{0.1861} \\
\textcolor{black}{QuantSR~\cite{qin2024quantsr}} & \textcolor{black}{$\times$4} & \textcolor{black}{2/2} & \textcolor{black}{17.99} & \textcolor{black}{0.3127} & \textcolor{black}{0.8593} & \textcolor{black}{0.4102} & \textcolor{black}{193.29} & \textcolor{black}{8.2745} & \textcolor{black}{40.08} & \textcolor{black}{0.3536} & \textcolor{black}{0.1861} \\
\rowcolor{colorTabTop}
\textcolor{black}{QuantSR+} & \textcolor{black}{$\times$4} & \textcolor{black}{2/2} & \textcolor{black}{18.12} & \textcolor{black}{0.3155} & \textcolor{black}{0.8362} & \textcolor{black}{0.3965} & \textcolor{black}{176.50} & \textcolor{black}{8.1393} & \textcolor{black}{40.81} & \textcolor{black}{0.3669} & \textcolor{black}{0.1719} \\
\hline
\end{tabular}
}
\caption{\textcolor{black}{Quantitative comparison on diffusion-based SR backbones under quantization on DIV2K-Val.}}
\label{tab:diffusion_quant_compare_div2kval}
\end{table*}

\begin{table*}[!t]
\small
\centering
\resizebox{0.99\textwidth}{!}{
\begin{tabular}{lcccccccccc}
\hline
\rowcolor{colorTabTop}
Method &
Body &
H/T &
Factor &
Params (K) &
Ops (G) &
Set5 (PSNR) &
\textcolor{black}{Hardware path} &
\textcolor{black}{Latency} &
\textcolor{black}{Bandwidth} &
\textcolor{black}{Throughput} \\
\hline \hline
SRResNet & 32/32 & 32/32 & - & 1,515 (0\%) & 90.1 (0\%) & 32.16 &
\textcolor{black}{ZCU104 PS, ARM Cortex-A53} &
\textcolor{black}{4.54 s} &
\textcolor{black}{23 MB/s} &
\textcolor{black}{0.22 FPS} \\
\hline
ODM~\cite{hong2024overcoming} & 4/4 & 32/32 & $l$ & 303 ($\downarrow$ 80.0\%) & 20.2 ($\downarrow$ 77.5\%) & 32.00 & & \textcolor{black}{2.15 s} & \textcolor{black}{21 MB/s} &
\textcolor{black}{0.47 FPS} \\
QuantSR~\cite{qin2024quantsr} & 4/4 & 32/32 & $c$ & 303 ($\downarrow$ 80.0\%) & 20.2 ($\downarrow$ 77.5\%) & 32.00 &
\textcolor{black}{Body on PL + Head/Tail on PS} &
\textcolor{black}{2.15 s} &
\textcolor{black}{21 MB/s} &
\textcolor{black}{0.47 FPS} \\
\rowcolor{colorTabTop}
QuantSR+ & 4/4 & 32/32 & $c$ & 303 ($\downarrow$ 80.0\%) & 20.2 ($\downarrow$ 77.5\%) & 32.13 & & \textcolor{black}{2.15 s} & \textcolor{black}{21 MB/s} &
\textcolor{black}{0.47 FPS} \\
\hline
\rowcolor{colorTabTop}
\textcolor{black}{QuantSR+} & \textcolor{black}{4/4} & \textcolor{black}{8/8} & \textcolor{black}{$c$} & \textcolor{black}{248 ($\downarrow$ 83.6\%)} &
\textcolor{black}{15.0 ($\downarrow$ 83.4\%)} &
\textcolor{black}{32.02} &
\textcolor{black}{Body on PL + Head/Tail on PL} &
\textcolor{black}{33 ms} &
\textcolor{black}{6.2 MB/s} &
\textcolor{black}{30.3 FPS} \\
\hline
\hline
ODM~\cite{hong2024overcoming} & 2/2 & 32/32 & $l$ & 161 ($\downarrow$ 89.4\%) & 10.9 ($\downarrow$ 87.9\%) & 31.81 & & \textcolor{black}{2.13 s} &
\textcolor{black}{21 MB/s} &
\textcolor{black}{0.47 FPS} \\
QuantSR~\cite{qin2024quantsr} & 2/2 & 32/32 & $c$ & 161 ($\downarrow$ 89.4\%) & 10.9 ($\downarrow$ 87.9\%) & 31.30 &
\textcolor{black}{Body on PL + Head/Tail on PS} &
\textcolor{black}{2.13 s} &
\textcolor{black}{21 MB/s} &
\textcolor{black}{0.47 FPS} \\
\rowcolor{colorTabTop}
QuantSR+ & 2/2 & 32/32 & $c$ & 161 ($\downarrow$ 89.4\%) & 10.9 ($\downarrow$ 87.9\%) & 31.89 & & \textcolor{black}{2.13 s} &
\textcolor{black}{21 MB/s} &
\textcolor{black}{0.47 FPS} \\
\hline
\rowcolor{colorTabTop}
\textcolor{black}{QuantSR+} & \textcolor{black}{2/2} & \textcolor{black}{8/8} & \textcolor{black}{$c$} &
\textcolor{black}{108 ($\downarrow$ 92.9\%)} &
\textcolor{black}{5.8 ($\downarrow$ 93.7\%)} &
\textcolor{black}{31.76} &
\textcolor{black}{Body on PL + Head/Tail on PL} &
\textcolor{black}{24 ms} &
\textcolor{black}{8.5 MB/s} &
\textcolor{black}{41.7 FPS} \\
\hline \hline
\end{tabular}
}
\caption{Efficiency comparison on SRResNet ($\times$4, 16 residual blocks). The input size for calculating theoretical Ops is $3\times256\times256$. ``Factor'' denotes the scaling-factor granularity, where $c$ and $l$ indicate channel-wise and layer-wise scaling, respectively. \textcolor{black}{Hardware metrics are evaluated on the AMD/Xilinx ZCU104 board with a ZU7EV FPGA using an LR input size of $3\times64\times64$ and an HR output size of $3\times256\times256$. PS and PL denote the processing system and programmable logic on the ZCU104 board, respectively. Bandwidth denotes the effective average off-chip DDR memory bandwidth, and throughput denotes the end-to-end frame processing rate measured on the complete deployment path. The rows with FP32 head/tail (H/T) follow the original QuantSR+ protocol, while the rows with INT8 head/tail correspond to a deployment-oriented, fully-integer FINN-style implementation.}}
\label{tab:frb_comp_ratio_speedup_grouped}
\end{table*}

\section{Experiment Results}

We follow the standard procedure for image SR by training on the DIV2K dataset~\cite{timofte2017ntire} and evaluating on commonly used benchmark datasets, including Set5~\cite{bevilacqua2012low}, Set14~\cite{zeyde2012single}, B100~\cite{martin2001database}, Urban100~\cite{huang2015single}, and Manga109~\cite{matsui2017sketch}.
We report the reconstruction accuracy of the comparison methods and our proposed QuantSR+ in terms of PSNR and SSIM~\cite{wang2004image} on the Y channel of the YCbCr color space. Additionally, we present both the savings of the storage usage and the computational complexity of the quantized SR models compared to their full-precision counterparts.

We apply our proposed quantization method to image SR models, including convolution-based and transformer-based SR architectures. \textcolor{black}{For convolution-based models, we follow the approaches in DoReFa~\cite{zhou2016dorefa}, PAMS~\cite{li2020pams}, CADyQ~\cite{hong2022cadyq}, QuantSR~\cite{qin2024quantsr}, AdaBM~\cite{hong2024adabm}, Granular-DQ~\cite{wang2025thinking}, CABM~\cite{tian2023cabm}, I$^2$NQ~\cite{sun20242}, RefQSR~\cite{lee2024refqsr}, PTQ4SR~\cite{tu2023toward}, DAQ~\cite{hong2022daq}, and ODM~\cite{hong2024overcoming}, and we use the SRResNet backbone to construct the model. For transformer-based SR models, we quantize the lightweight version of SwinIR~\cite{liang2021swinir}, comparing our QuantSR+ with QuantSR~\cite{qin2024quantsr}, ODM~\cite{hong2024overcoming}, and 2DQuant~\cite{liu20252dquant}.}
\textcolor{black}{We also apply our method to diffusion-based SR backbones~\cite{wang2024exploiting}.
The implementation of these methods follows their papers~\cite{zhou2016dorefa,sun20242} and released codes~\cite{cadyq_github,pams_github,quantsr_github}, or follows results reported in their original papers directly~\cite{wang2025thinking,tian2023cabm,hong2024overcoming}.} All implemented models are trained under identical settings and pipelines to ensure consistency. We quantize the weights and activations in the body part of each model to 2, 4, or 8 bit-widths. We denote the $w$-bit weight and $a$-bit activation as $w$/$a$. For a fair comparison, QuantSR reports performance for variants with the same depth as the other networks, \textit{i.e.}, 16 blocks for SRResNet and 4 blocks for SwinIR.

During the training process, we adopt the data augmentation techniques used in prior studies~\cite{lim2017enhanced,zhang2018learning,xin2020binarized,liang2021swinir,qin2024quantsr}, including random rotations of 90$^{\circ}$, 180$^{\circ}$, and 270$^{\circ}$, as well as horizontal flipping.
The quantized SR models are trained for 300K iterations, with each training batch comprising 32 image patches, each with an input size of 64$\times$64. We optimize our model using the Adam optimizer~\cite{kingma2014adam}, starting with an initial learning rate of 2e-4, which is halved after the 250K-th iteration.
\textcolor{black}{Notably, on SwinIR's W2A2, we aligned the training settings with the ODM~\cite{hong2024overcoming}, \textit{i.e.}, 300K iterations with an initial learning rate of 1e-4, halved every 75K iterations, and we further present the comparative analysis of different strategies based on this case in Section~6.6.}

\begin{table*}[t]
\footnotesize
\centering
\begin{center}
\resizebox{1\columnwidth}{!}{
\begin{tabular}{lcccccccccccc}
\hline
\rowcolor{colorTabTop}
& Scale & {\#Bit} & \multicolumn{2}{c}{Set5} & \multicolumn{2}{c}{Set14} & \multicolumn{2}{c}{B100} & \multicolumn{2}{c}{Urban100} & \multicolumn{2}{c}{Manga109} \\
\cline{4-13}
\rowcolor{colorTabTop}
\multirow{-2}{*}{Method} &  & ($w$/$a$) & PSNR & SSIM & PSNR & SSIM & PSNR & SSIM & PSNR & SSIM & PSNR & SSIM \\
\hline
\hline
SRResNet & $\times$4 & 32/32 & 32.16 & 0.8951 & 28.60 & 0.7822 & 27.58 & 0.7364 & 26.11 & 0.7870 & 30.46 & 0.9089 \\
\hline
\textcolor{black}{Vanilla (DoReFa)} & \textcolor{black}{$\times$4} & \textcolor{black}{2/2} & \textcolor{black}{30.54} & \textcolor{black}{0.8610} & \textcolor{black}{27.50} & \textcolor{black}{0.7538} & \textcolor{black}{26.90} & \textcolor{black}{0.7098} & \textcolor{black}{24.44} & \textcolor{black}{0.7242} & \textcolor{black}{27.31} & \textcolor{black}{0.8502} \\
\textcolor{black}{RBD} & \textcolor{black}{$\times$4} & \textcolor{black}{2/2} & \textcolor{black}{30.75} & \textcolor{black}{0.8686} & \textcolor{black}{27.78} & \textcolor{black}{0.7514} & \textcolor{black}{26.99} & \textcolor{black}{0.7034} & \textcolor{black}{25.06} & \textcolor{black}{0.7394} & \textcolor{black}{28.46} & \textcolor{black}{0.8722} \\
\textcolor{black}{QSA} & \textcolor{black}{$\times$4} & \textcolor{black}{2/2} & \textcolor{black}{31.01} & \textcolor{black}{0.8738} & \textcolor{black}{27.95} & \textcolor{black}{0.7573} & \textcolor{black}{27.11} & \textcolor{black}{0.7095} & \textcolor{black}{25.25} & \textcolor{black}{0.7476} & \textcolor{black}{28.80} & \textcolor{black}{0.8790} \\
\textcolor{black}{SFD} & \textcolor{black}{$\times$4} & \textcolor{black}{2/2} & \textcolor{black}{31.21} & \textcolor{black}{0.8776} & \textcolor{black}{28.07} & \textcolor{black}{0.7616} & \textcolor{black}{27.20} & \textcolor{black}{0.7139} & \textcolor{black}{25.39} & \textcolor{black}{0.7538} & \textcolor{black}{29.07} & \textcolor{black}{0.8840} \\
\textcolor{black}{RBD+QSA} & \textcolor{black}{$\times$4} & \textcolor{black}{2/2} & \textcolor{black}{31.45} & \textcolor{black}{0.8819} & \textcolor{black}{28.22} & \textcolor{black}{0.7667} & \textcolor{black}{27.31} & \textcolor{black}{0.7192} & \textcolor{black}{25.51} & \textcolor{black}{0.7616} & \textcolor{black}{29.41} & \textcolor{black}{0.8901} \\
\rowcolor{colorTabTop}
\textcolor{black}{QuantSR+} & \textcolor{black}{$\times$4} & \textcolor{black}{2/2} & \textcolor{black}{{31.89}} & \textcolor{black}{{0.8908}} & \textcolor{black}{{28.37}} & \textcolor{black}{{0.7763}} & \textcolor{black}{{27.43}} & \textcolor{black}{{0.7308}} & \textcolor{black}{{25.58}} & \textcolor{black}{{0.7702}} & \textcolor{black}{{29.67}} & \textcolor{black}{{0.8974}} \\
\hline
\end{tabular}}
\caption{\textcolor{black}{Ablation results ($\times$4 scale) for Redistribution-driven Bit Determination (RBD), Quantized Slimmable Architecture (QSA), and Slimming-guided Function-localized Distillation (SFD) in QuantSR+ under the 2-bit setting.}}
\label{tab:frb_ablation_dorefa_rw_kd}
\end{center}
\end{table*}

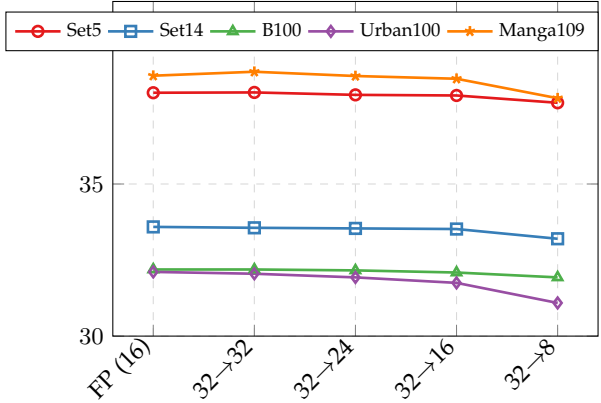
\begin{figure}[t]
\centering
\begin{tikzpicture}
    \begin{groupplot}[
        group style={
            vertical sep=1.cm,
            xlabels at=edge bottom,
            ylabels at=edge left
        },
        width=0.9\textwidth,
        height=6cm,
        x tick label style={rotate=45, anchor=east},
        xtick={1,2,3,4,5},
        xticklabels={FP (16), 32$\rightarrow$32, 32$\rightarrow$24, 32$\rightarrow$16, 32$\rightarrow$8},
        grid=major,
        grid style={dashed, gray!30}
    ]

    \nextgroupplot[
        ymin=30, ymax=41,
        legend columns=5,
        legend pos=north east,
        legend style={xshift=0.2cm, yshift=0cm}
    ]
    \addplot[color=set5, mark=o, line width=1pt] coordinates {
        (1,38.00) (2,38.01) (3,37.93) (4,37.91) (5,37.67)
    };
    \addlegendentry{Set5}
    \addplot[color=set14, mark=square, line width=1pt] coordinates {
        (1,33.59) (2,33.56) (3,33.54) (4,33.52) (5,33.20)
    };
    \addlegendentry{Set14}
    \addplot[color=b100, mark=triangle, line width=1pt] coordinates {
        (1,32.19) (2,32.19) (3,32.16) (4,32.09) (5,31.93)
    };
    \addlegendentry{B100}
    \addplot[color=urban100, mark=diamond, line width=1pt] coordinates {
        (1,32.11) (2,32.05) (3,31.93) (4,31.75) (5,31.09)
    };
    \addlegendentry{Urban100}
    \addplot[color=manga109, mark=star, line width=1pt] coordinates {
        (1,38.56) (2,38.69) (3,38.55) (4,38.46) (5,37.82)
    };
    \addlegendentry{Manga109}
    \end{groupplot}
\end{tikzpicture}
\caption{\textcolor{black}{QSA ablation results (PSNR) for different block configurations (x2 Scale, 4/4-bit). 16 (FP) means a 16 block model with 32-bit; others are 4/4-bit models quantized by QuantSR+.}}
\label{fig:qsa_ablation}
\end{figure}

\begin{table}[th]
\small
\centering
\resizebox{1\columnwidth}{!}{
{\color{black}
\begin{tabular}{lcccccc}
\hline
\rowcolor{colorTabTop}
Method & Scale & \#Bit & \multicolumn{2}{c}{Set5} & \multicolumn{2}{c}{Set14} \\
\cline{4-7}
\rowcolor{colorTabTop}
& & ($w/a$) & PSNR & SSIM & PSNR & SSIM \\
\hline
\hline
SRResNet~\cite{ledig2017photo} & $\times$2 & 32/32 & 38.00 & 0.9605 & 33.59 & 0.9171 \\
\hline
QuantSR+ & $\times$2 & 1/1 & 36.66 & 0.9548 & 32.39 & 0.9069 \\
\hline
\hline
SRResNet~\cite{ledig2017photo} & $\times$4 & 32/32 & 32.16 & 0.8951 & 28.60 & 0.7822 \\
\hline
QuantSR+ & $\times$4 & 1/1 & 30.32 & 0.8604 & 27.40 & 0.7505 \\
\hline
\end{tabular}
}
}
\caption{\textcolor{black}{$1/1$-bit quantitative results on SRResNet.}}
\label{tab:quantsr_srresnet_1bit}
\end{table}

\subsection{Main Results}
\label{subsec:image_sr}
For our studies, we adopt two distinct architectures as the backbone models: SRResNet~\cite{ledig2017photo}, a convolutional neural network (CNN) with 1,367K parameters for $\times$2 upscaling and 1,515K parameters for $\times$4, and SwinIR\_S~\cite{liang2021swinir}, a compact transformer-based model featuring 910K and 930K parameters for $\times$2 and $\times$4 scales, respectively.
\textcolor{black}{The quantized versions of our QuantSR+ are evaluated alongside established quantization approaches, including DoReFa~\cite{zhou2016dorefa}, PAMS~\cite{li2020pams}, CADyQ~\cite{hong2022cadyq}, QuantSR~\cite{qin2024quantsr}, AdaBM~\cite{hong2024adabm}, Granular-DQ~\cite{wang2025thinking}, CABM~\cite{tian2023cabm}, I$^2$NQ~\cite{sun20242}, ODM~\cite{hong2024overcoming}, RefQSR~\cite{lee2024refqsr}, PTQ4SR~\cite{tu2023toward}, DAQ~\cite{hong2022daq}, and 2DQuant~\cite{liu20252dquant}.}

\textcolor{black}{For the convolution-based SRResNet, Table~\ref{tab:quantsr_srresnet} reports quantitative results at $\times 2$ and $\times 4$ under 2-bit and 4-bit quantization, with additional 8-bit references available at $\times 2$. In the ultra-low 2-bit setting, QuantSR+ consistently improves over the strongest prior fixed-precision baselines on all five benchmarks. \textcolor{black}{On Set5, it improves PSNR/SSIM from 31.81/0.8880 (ODM) to 31.89/0.8908 at $\times 4$, corresponding to gains of 0.08dB/0.0028.} Similar gains are also observed on Set14, B100, Urban100, and Manga109, demonstrating the robustness of QuantSR+ across different image contents and scaling factors. In the 4-bit setting, QuantSR+ remains highly competitive, while its enhanced variant QuantSR+$^{*}$, obtained by incorporating the activation shifting and partition strategy from I$^2$NQ~\cite{sun20242}, achieves the best overall performance. For example, QuantSR+$^{*}$ reaches 32.17dB/0.8951 at $\times 4$, even surpassing the full-precision SRResNet (32.16dB/0.8951 at $\times 4$). Compared with mixed-precision methods~\cite{hong2024adabm,tian2023cabm,wang2025thinking}, QuantSR+$^{*}$ also shows clear advantages, e.g., exceeding Granular-DQ by 0.03dB on Urban100 at $\times 4$. These results demonstrate both the effectiveness of QuantSR+ in low-bit SR and its compatibility with other advanced techniques.}

As for transformer-based models, our QuantSR+ approach effectively mitigates the accuracy gap caused by quantization (TABLE~\ref{tab:quantsr_swinir}). For example, with the 2-bit quantization, QuantSR+ outperforms QuantSR~\cite{qin2024quantsr} by a substantial margin in $\times$4 scale, achieving gains of 0.17 dB in PSNR and 0.0015 in SSIM on Set5.
\textcolor{black}{We also compare with recent quantization methods that natively support the SR transformer, and our QuantSR+ significantly improves the accuracy, \textit{e.g.}, 1.81 dB PSNR over 2DQuant~\cite{liu20252dquant} and 0.29 dB over ODM~\cite{hong2024overcoming} on Urban100 at $\times$4 scale under the same 300K iterations.}

\textcolor{black}{We further verify the generalization ability of our QuantSR+ on diffusion-based SR backbones in Table~\ref{tab:diffusion_quant_compare_div2kval}. For example, 4-bit QuantSR+ achieves the best PSNR, SSIM, LPIPS, DISTS, and FID among all quantized baselines. Compared with QuantSR, it improves PSNR and SSIM by 0.06 dB and 0.0148, while reducing LPIPS, DISTS, and FID by 0.0238, 0.0118, and 9.62.}

\begin{figure}[t]
\centering
\includegraphics[width=\textwidth]{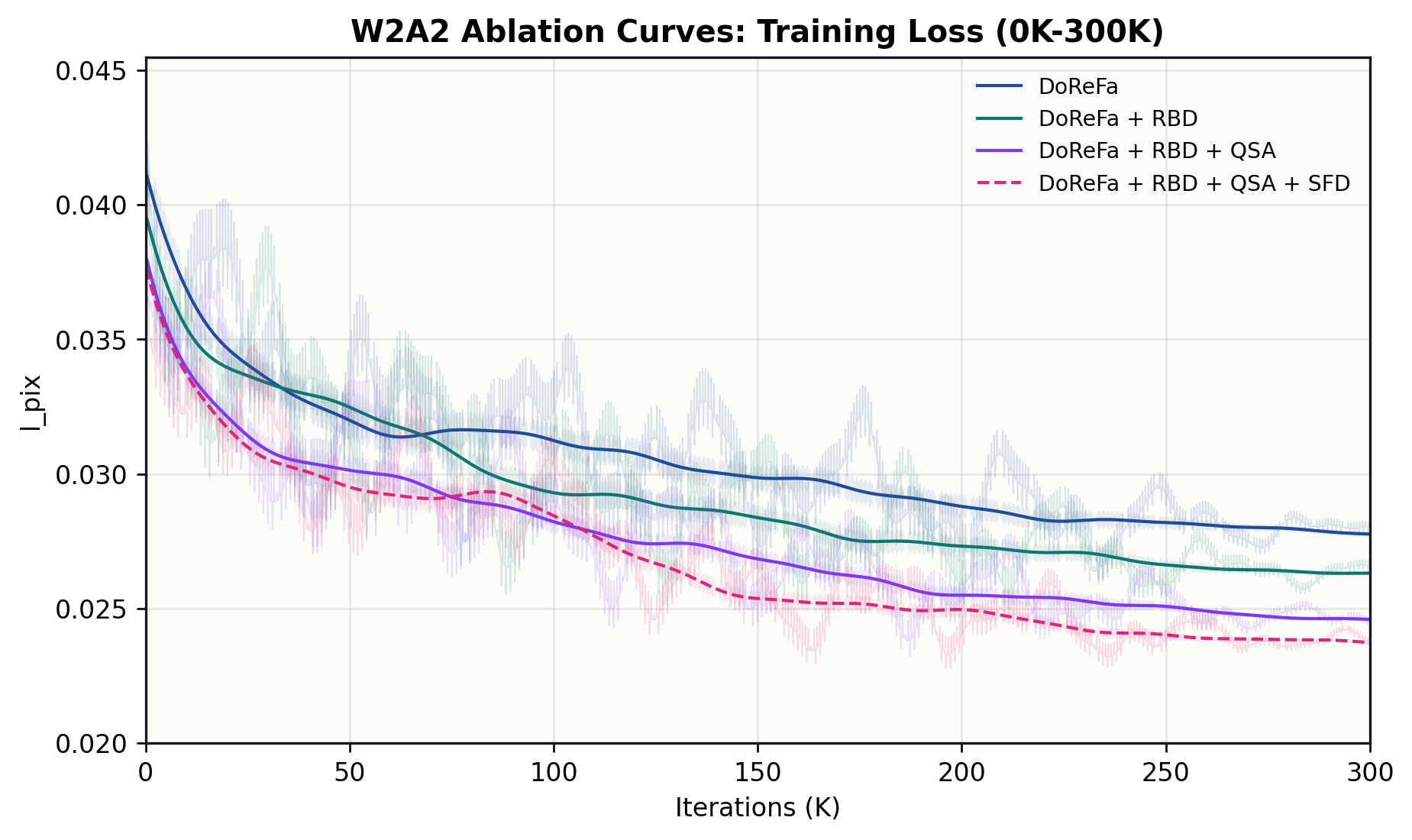}
\caption{\textcolor{black}{W2A2 ablation curves of training loss on SRResNet $\times4$. The translucent curves denote the original per-iteration losses, and the solid/dashed curves denote the smoothed trends. Lower and smoother trajectories indicate more stable optimization.}}
\label{fig:sfd_loss_curve}
\end{figure}

\begin{figure*}[t]
\scriptsize
\centering
\begin{adjustbox}{valign=t}
\begin{tabular}{c}
\includegraphics[width=\textwidth]{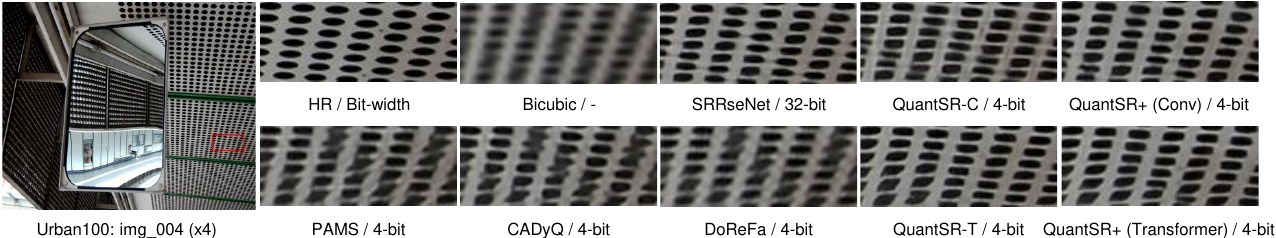} \\
\end{tabular}
\end{adjustbox}
\caption{Visualization ($\times$4) for quantized SR models in terms of 4-bit setting.}
\label{fig:quantsr_visual_result_SRBIX4_lightweight}
\end{figure*}

\begin{figure*}[tbp]
\centering
\subfloat[$\hat{v}_b$ for weight quantizer.]{\includegraphics[width=.32\linewidth]{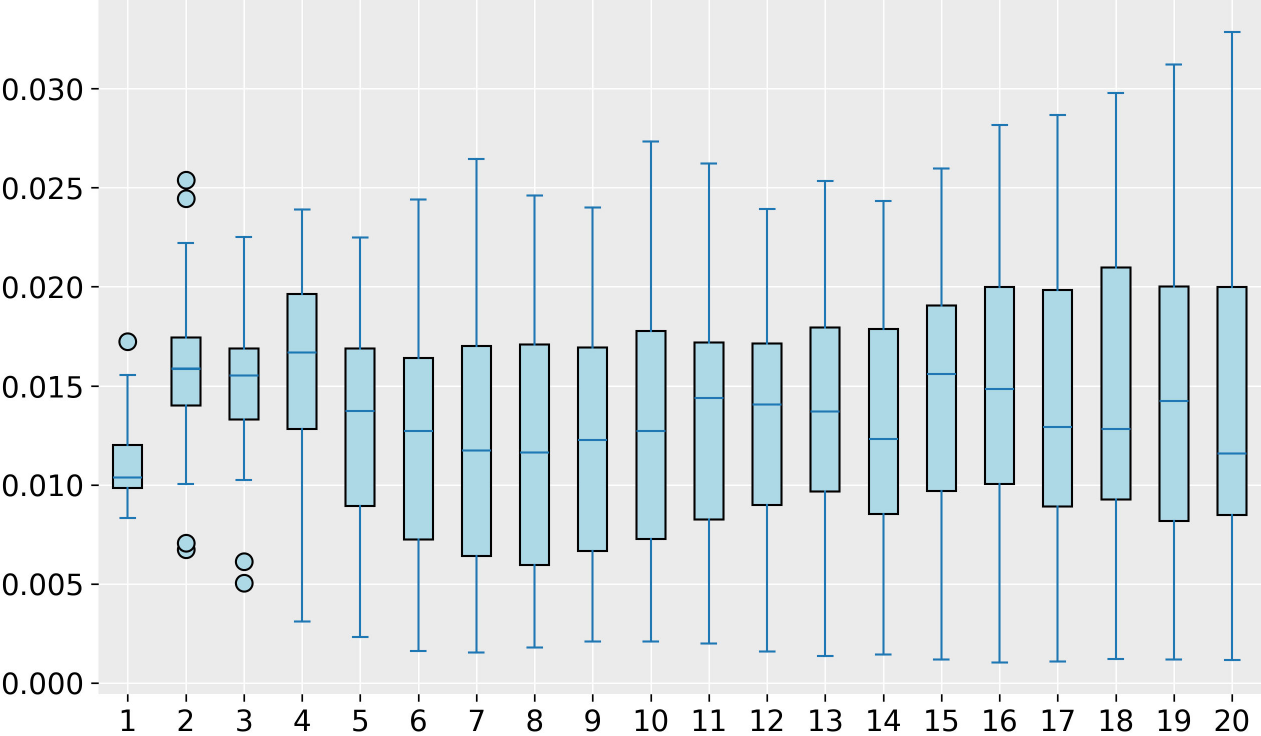}}\hspace{5pt}
\subfloat[$\hat{\tau}$ for activation quantizer.]{\includegraphics[width=.32\linewidth]{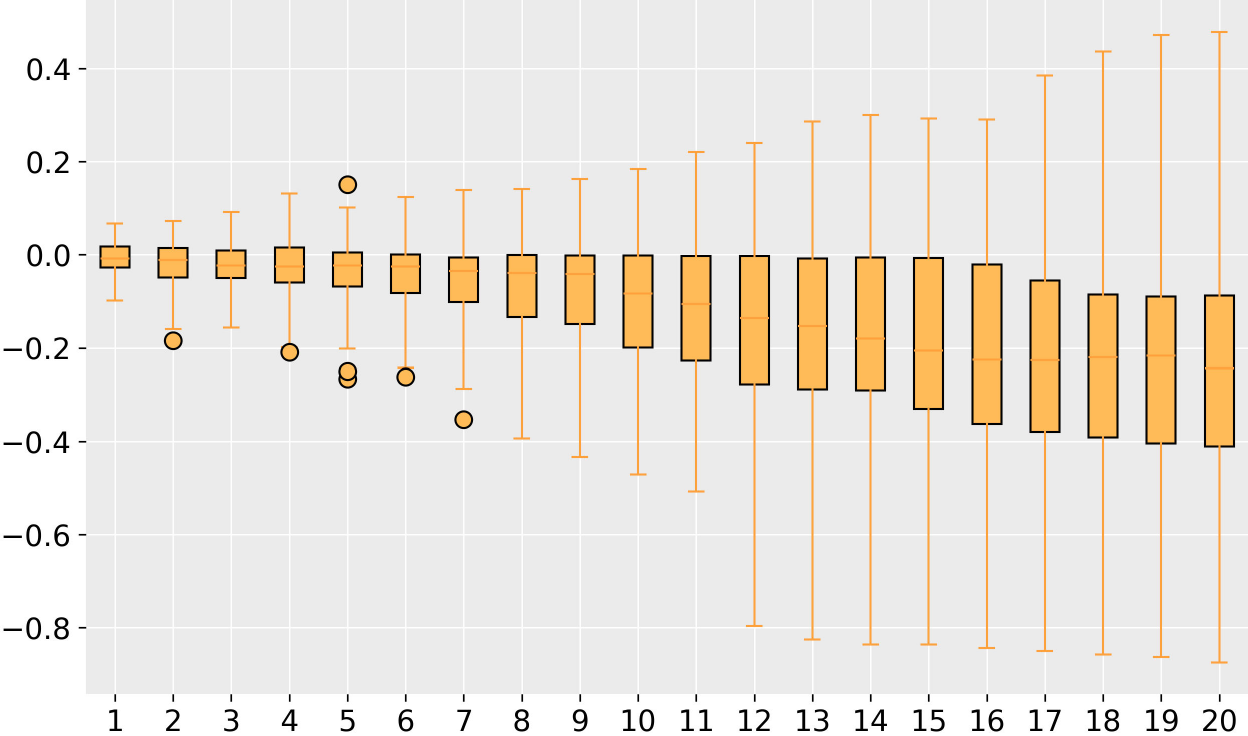}}\hspace{5pt}
\subfloat[Gradient effect of $\phi(\cdot)$]{\includegraphics[width=.32\linewidth]{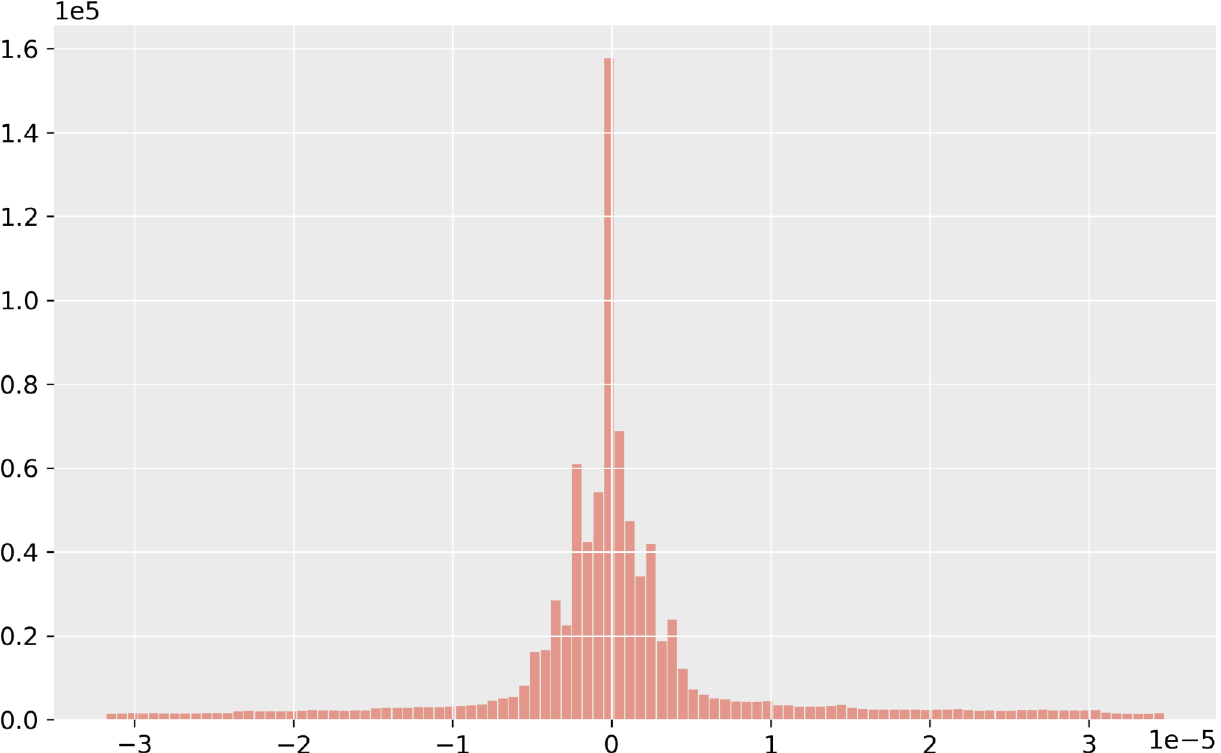}}
\caption{The statistics of operator parameters in QuantSR+. The visualizations demonstrate the improvement in representation from quantized operators in forward and backward propagation.}
\label{fig:method1-row}
\end{figure*}

\textcolor{black}{By integrating quantization with a dynamically optimized lightweight architecture, QuantSR+ achieves substantial compression while maintaining strong reconstruction accuracy. TABLE~\ref{tab:frb_comp_ratio_speedup_grouped} reports both theoretical efficiency and practical hardware deployment results. Following PAMS~\cite{li2020pams}, the original QuantSR+ protocol quantizes the high-level feature extraction layers, reducing Params/Ops by 80.0\%/77.5\% at 4-bit and 89.4\%/87.9\% at 2-bit under the theoretical complexity protocol with a $3\times256\times256$ input size, while outperforming representative layer-wise and channel-wise methods, i.e., ODM~\cite{hong2024overcoming} and QuantSR~\cite{qin2024quantsr}. Since the backbone and bit-width mainly determine Params and Ops, the original-protocol rows for ODM, QuantSR, and QuantSR+ share the same Params/Ops, and the resulting accuracy gains mainly stem from improved quantizer design, scaling strategy, and reconstruction capability. We further evaluate QuantSR+ on the AMD/Xilinx ZCU104 board with a ZU7EV FPGA, using an LR input size of $3\times64\times64$ and an HR output size of $3\times256\times256$. With FP32 head/tail and a low-bit body, QuantSR+ achieves about 2.1$\times$ end-to-end speedup and 0.47 FPS throughput. With INT8 head/tail and a fully programmable-logic FINN-style dataflow implementation, the deployment-oriented W4A4 and W2A2 QuantSR+ implementations achieve 33 ms and 24 ms latency, corresponding to 138$\times$ and 189$\times$ system-level speedup, respectively. The corresponding end-to-end throughput reaches 30.3 FPS and 41.7 FPS, with effective average off-chip bandwidths of 6.2 MB/s and 8.5 MB/s. More detailed efficiency analysis is provided in Section~6.1 and Section~6.2.}

\subsection{Ablation Results}
To demonstrate the effectiveness of the techniques in QuantSR+, we perform detailed ablation studies on RBD, QSA, and SFD. Our study is based on the SRResNet~\cite{ledig2017photo} backbone network, and uses the DoReFa quantization method as the vanilla quantized SR baseline model~\cite{zhou2016dorefa}. We then construct ablation studies for each technique based on the vanilla method. \textcolor{black}{The experimental results are shown in TABLE~\ref{tab:frb_ablation_dorefa_rw_kd}, where we report the PSNR/SSIM values on five benchmark datasets.}

\textcolor{black}{As shown in TABLE~\ref{tab:frb_ablation_dorefa_rw_kd}, the effectiveness of RBD, QSA, and SFD is evaluated separately under the 2-bit W2A2 setting.}
\textcolor{black}{Compared with the vanilla DoReFa model, RBD improves PSNR on all five benchmarks by 0.09-1.15 dB; for SSIM, it brings clear gains on Set5, Urban100, and Manga109, while the Set14 and B100 SSIM values are slightly lower than the vanilla baseline by 0.0024 and 0.0064, respectively.}
\textcolor{black}{QSA also yields stable PSNR gains of 0.21-1.49 dB, while its SSIM improves by 0.0035-0.0288 across four benchmarks and shows only a marginal 0.0003 drop on B100, demonstrating that the slimmer architecture largely preserves reconstruction quality while reducing computational and memory costs.}
We further investigate the impact of different block configurations under QSA (Fig.~\ref{fig:qsa_ablation}): our 4/4-bit quantized models maintain consistent performance even as we gradually reduce the block count from 32 to 24, then to 16, and finally to 8. Specifically, on Set5, the PSNR only slightly drops from 38.01 dB (32$\rightarrow$32) to 37.67 dB (32$\rightarrow$8). In particular, the 16-block configuration adapted by our QuantSR+ strikes a favorable balance between performance preservation and computational efficiency. This validates that QSA effectively preserves reconstruction quality while achieving significant model compression. \textcolor{black}{Meanwhile, SFD further boosts reconstruction quality without changing the inference topology, and achieves the best results among the three individual variants, with improvements of 0.30-1.76 dB in PSNR and 0.0041-0.0338 in SSIM over the vanilla baseline.
When all components are combined, QuantSR+ further improves over the vanilla baseline by 0.53-2.36 dB in PSNR and 0.0210-0.0472 in SSIM, confirming that the three components make effective and complementary contributions.
We put more detailed results in Section~6.3.}

\textcolor{black}{Fig.~\ref{fig:sfd_loss_curve} provides empirical evidence for the improved optimization stability claimed in the method section. Under the challenging W2A2 setting, the vanilla DoReFa baseline exhibits a higher and more fluctuating loss trajectory, while each proposed component progressively improves the convergence behavior. The resulting full QuantSR+ variant achieves the lowest and smoothest loss curve, with a clear advantage from about 50K to 300K iterations. These observations are consistent with our design motivation that QuantSR+ improves stability jointly from the operator, architecture, and optimization levels. More detailed convergence analyses are provided in Section~6.4.}

\textcolor{black}{We also present the model performance at ultra-low 1-bit (TABLE~\ref{tab:quantsr_srresnet_1bit}). Our specially designed RBD for multi-bit fine-grained optimization struggles to perform effectively at 1-bit because it cannot fully leverage cooperation among multiple bits to optimize performance. Therefore, the co-design of quantizer formulas, training strategies, and network architectures specifically for a single bit is a promising direction for future research.}

\subsection{Visualization Results}
\label{subsec:visualization}

Fig.~\ref{fig:quantsr_visual_result_SRBIX4_lightweight} presents more qualitative results of different methods at a $\times$4 scale, focusing on 4-bit quantization. Our method, QuantSR+, is evaluated against existing quantization approaches, including DoReFa~\cite{zhou2016dorefa}, PAMS~\cite{li2020pams}, CADyQ~\cite{hong2022cadyq}, and QuantSR~\cite{qin2024quantsr}. Both SRResNet- and SwinIR-based quantized variants of QuantSR and QuantSR+ are included in the comparison. Our results demonstrate that QuantSR+ is more effective in preserving structural details and minimizing blurring artifacts compared to other quantization techniques. Notably, the visual gap between our quantized model and the full-precision version is minimal, reinforcing the quantitative findings in TABLE~\ref{tab:quantsr_srresnet} and TABLE~\ref{tab:quantsr_swinir} and further validating the effectiveness of QuantSR+.

Fig.~\ref{fig:method1-row} provides a statistical analysis of the learnable parameters and functional behavior of our RBD throughout the training process, recorded at every 100 iterations. We analyze key variables in our quantization functions, including $\hat{v}_b$ from the activation quantizer, $\hat{\tau}$ from the weight quantizer, and the gradient effect of the activation transformation function $\phi(\cdot)$. Fig.~\ref{fig:method1-row} (a) and (b) reveal that these learnable parameters start with similar values but gradually diverge over training, indicating that the QuantSR+ quantizer becomes more adaptive and effectively compensates for information loss due to discretization.
In Fig.~\ref{fig:method1-row} (c), we examine the gradient behavior induced by the transformation function $\phi(\cdot)$ during back-propagation. While the forward propagation results and training stability remain unaffected, the gradient adjustments introduced by this function play a critical role in guiding optimization. This contribution significantly enhances the training dynamics of QuantSR+, leading to improved overall performance.

\section{Conclusion}
This paper introduces QuantSR+, a novel quantized image SR network designed to address the accuracy and efficiency challenges in low-bit SR models. Our proposed network features three key innovations: Redistribution-driven Bit Determination (RBD), which enhances the representational capability of quantization operators; Quantized Slimmable Architecture (QSA), which enables progressive structural evolution to push accuracy beyond full-precision limits; and Slimming-guided Function-localized Distillation (SFD), which mitigates quantization-induced errors and accelerates convergence. We demonstrate that QuantSR+ surpasses existing quantized SR networks and general quantization methods with SOTA results across various bit widths and evaluation benchmarks, and achieves up to 88\% computation and storage savings at 2-bit highlighting its computational efficiency. Finally, QuantSR+ shows comprehensive accuracy and efficiency gains across both convolutional and transformer-based architectures, demonstrating its versatility and broad applicability. The QuantSR+ framework opens up new possibilities for the integration of advanced image SR networks via low-bit quantization in real-time scenarios, such as resource-constrained mobile and embedded devices.

\ifCLASSOPTIONcompsoc
  \section*{Acknowledgments}
\else
  \section*{Acknowledgment}
\fi

This work is supported by the National Natural Science Foundation of China (62501386), CCF-Tencent Rhino-Bird Open Research Fund, CAAI-Tencent Rhino-Bird Open Research Fund, and the Swiss National Science Foundation (SNSF) project 200021E\_219943 Neuromorphic Attention Models for Event Data (NAMED). This work is supported by the fundamental research funds for the central universities, sponsored by the AI Hundred Schools Program, and carried out using the Ascend AI technology stack.

{
\bibliography{egbib}
\bibliographystyle{ieee_fullname}
}

\clearpage

\begin{IEEEbiography}[{\includegraphics[width=1in,height=1.25in,clip,keepaspectratio]{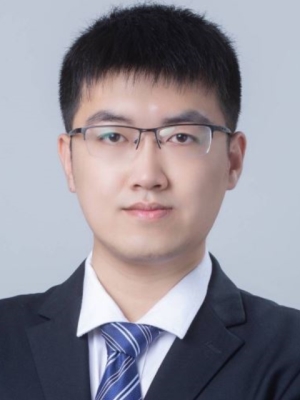}}]{Haotong Qin} is a Postdoctoral Researcher at the Center for Project-Based Learning (PBL) D-ITET at ETH Zürich. He was a Ph.D. student in the State Key Laboratory of Complex and Critical Software Environment at Beihang University. He obtained a B.Eng degree in computer science and engineering from Beihang University. His research interests include model compression and deployment toward efficient deep learning in real-world scenarios. He has published 35 papers in top-tier journals and conferences, such as IEEE TPAMI, IEEE TNNLS, IJCV, ICML, NeurIPS, ICLR, and CVPR. He serves as Guest Editor for Neural Networks, Area Chair in BMVC, and Reviewer for IEEE TPAMI, IEEE TIP, IEEE TNNLS, CVPR, etc.
\end{IEEEbiography}

\begin{IEEEbiography}[{\includegraphics[width=1in,height=1.25in,clip,keepaspectratio]{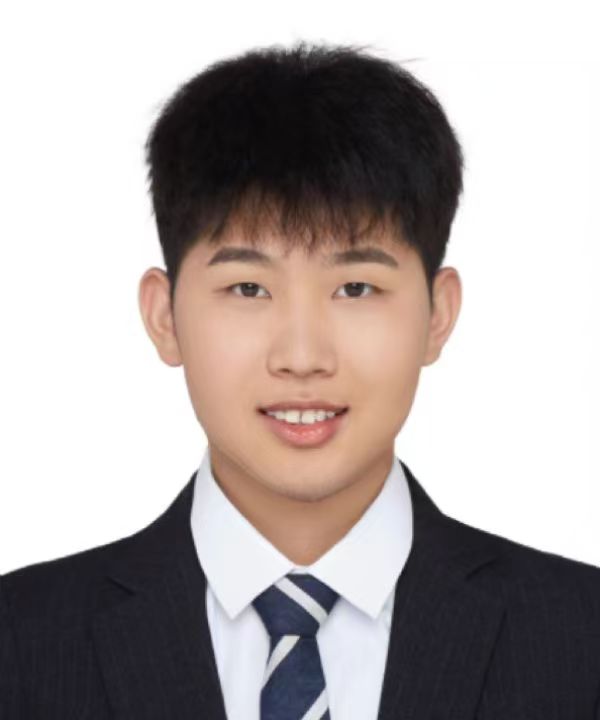}}]{Xudong Ma} received the B.E. degree in Network Engineering from the University of Electronic Science and Technology of China. He is currently working toward a PhD degree in the State Key Laboratory of Complex and Critical Software Environment at Beihang University under the supervision of Prof. Jie Luo. His research interests include model compression, knowledge graphs, and language models. He has published papers in NeurIPS, ICML, IJCAI, and IEEE TNNLS.
\end{IEEEbiography}

\begin{IEEEbiography}[{\includegraphics[width=1in,height=1.25in,clip,keepaspectratio]{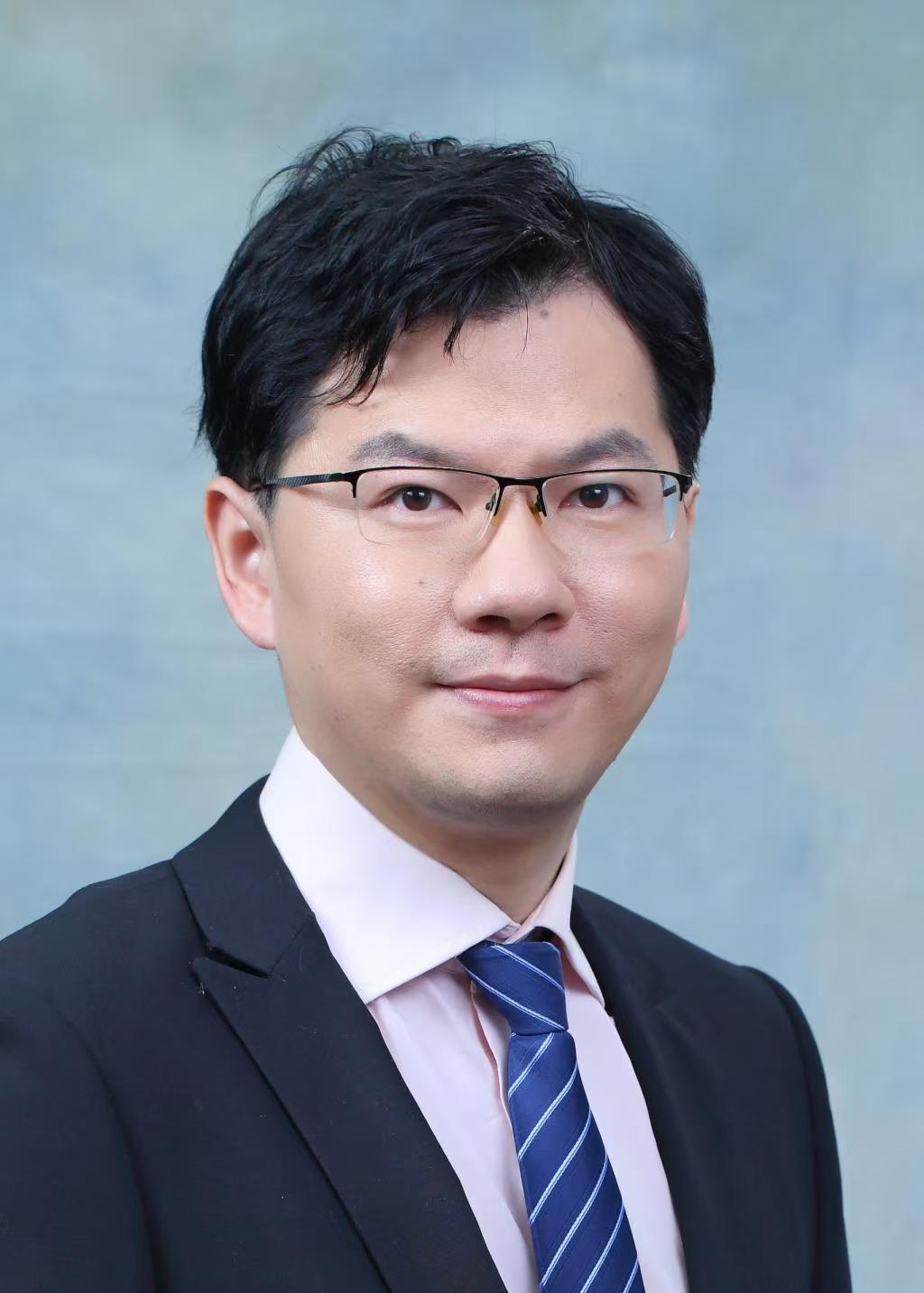}}]{Jie Luo} is now an Associate Professor in the State Key Laboratory of Complex and Critical Software Environment, School of Computer Science and Engineering, Beihang University. He received a B.S. degree from the School of Mathematical Sciences, Peking University, in 2003. He received a Ph.D. from Beihang University in 2012 under the supervision of Prof. Wei Li and visited the University of Washington as a joint PhD student. His research interests include logic foundations for computer science, knowledge engineering, and crowd intelligence. He has published over 40 papers in top conferences and journals in artificial intelligence and information security, such as NeurIPS, ICML, CVPR, IJCAI, BiBM, KSEM, and PR.
\end{IEEEbiography}

\begin{IEEEbiography}[{\includegraphics[width=1in,height=1.25in,clip,keepaspectratio]{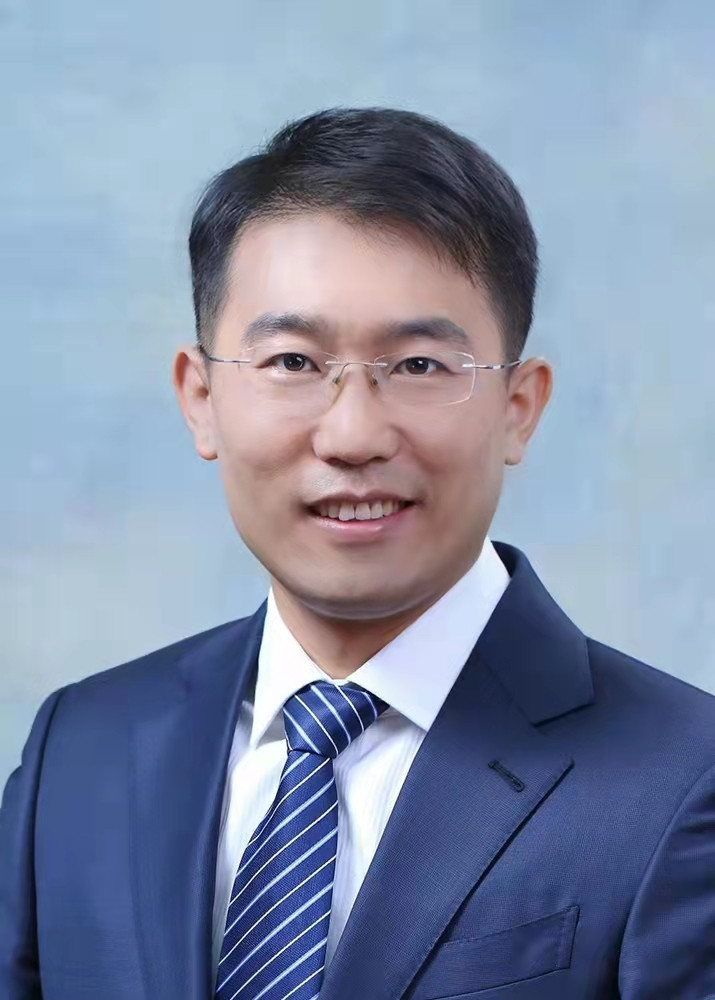}}]{Xianglong Liu} is currently a Professor, serves as the vice dean of the School of Computer Science and Engineering at Beihang University, and is also the deputy director of the State Key Laboratory of Complex and Critical Software Environment. He received his B.S. and Ph.D. degrees under the supervision of Prof. Wei Li and visited the DVMM Lab at Columbia University as a joint PhD student supervised by Prof. Shih-Fu Chang. He is the recipient of the China National Excellent Youth Science Fund. He has published over 100 papers in top conferences/journals in artificial intelligence and information security, such as NeurIPS, ICLR, CVPR, ICCV, CSS, and IJCV. He serves as Associate Editor and Guest for several SCI journals like Pattern Recognition and IET Image Processing, and as Promotion Editor for journals like Frontiers of Computer Science and Acta Aeronautica et Astronautica Sinica. He serves as Area Chair in top conferences such as AAAI and ACM MM and has frequently organized workshops and competitions in conferences like CVPR, IJCAI, and AAAI.
\end{IEEEbiography}

\begin{IEEEbiography}[{\includegraphics[width=1in,height=1.25in,clip,keepaspectratio]{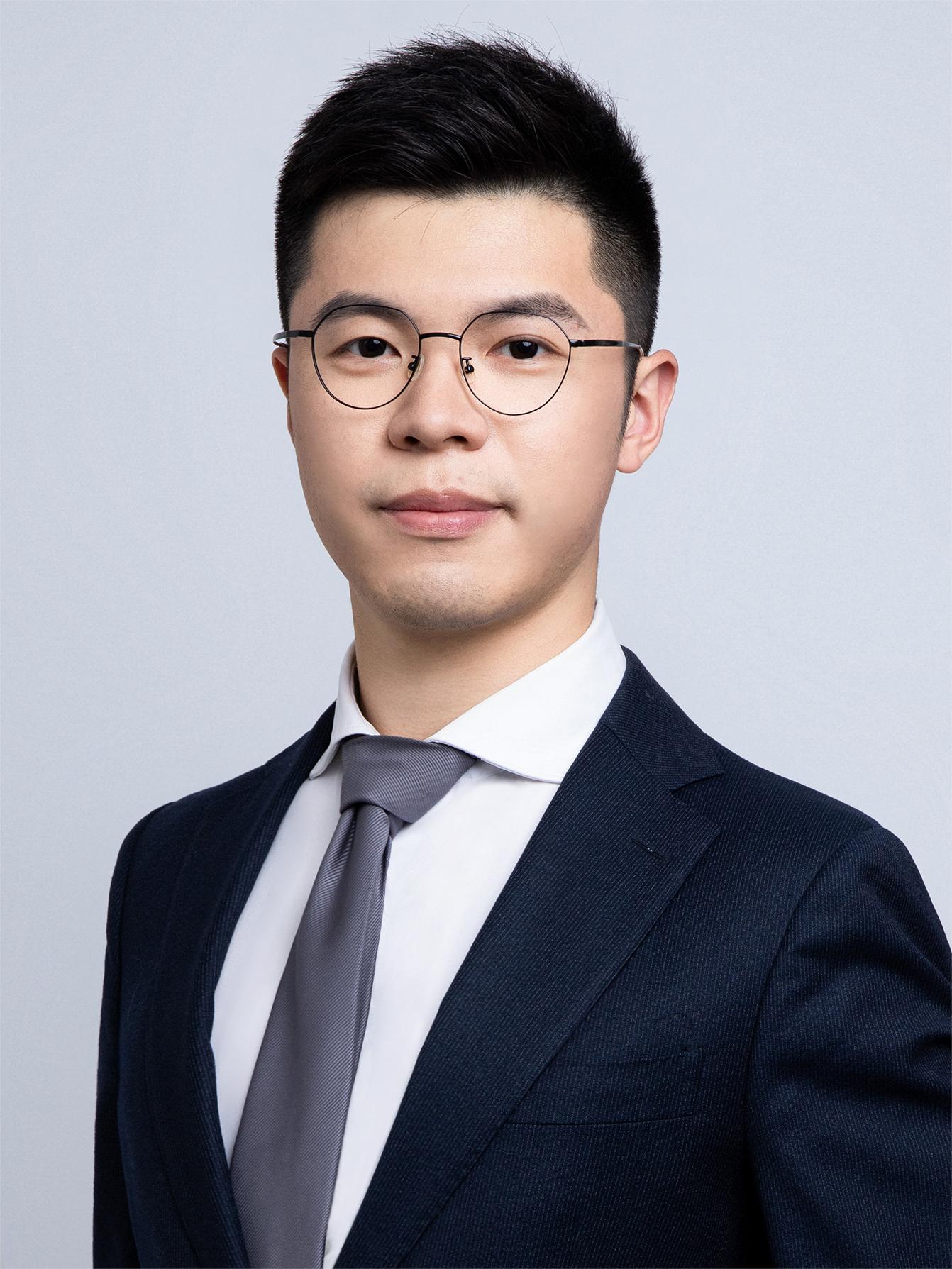}}]{Jinyang Guo} is an Assistant Professor at the State Key Laboratory of Complex \& Critical Software Environment, Institute of Artificial Intelligence, Beihang University, China. Previously, he obtained his B.Eng (Hons1) degree from the School of Electrical and Telecommunication, The University of New South Wales, Australia, and later received his Ph.D. from the School of Electrical and Information Engineering, The University of Sydney. His research interests include efficient and scalable AI computing (e.g., model-efficiency, data-efficiency, label-efficiency), neuromorphic computing (e.g., spiking neural networks), and AI4Science (e.g., AI for quantum computing).
\end{IEEEbiography}

\begin{IEEEbiography}[{\includegraphics[width=1in,height=1.25in,clip,keepaspectratio]{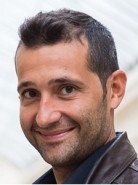}}]{Michele Magno} is currently a Privatdozent with the Department of Information Technology and Electrical Engineering (D-ITET), ETH Zürich, where he has been leading the D-ITET Center for project-based learning since 2020. He is also a Fellow of IEEE. He received master's and Ph.D. degrees in electronic engineering from the University of Bologna, Bologna, Italy, in 2004 and 2010, respectively. Since 2013, he has been with ETH Zürich, Zürich, Switzerland, and has become a Visiting Lecturer or a Professor at several universities, namely, the University of Nice Sophia, Nice, France; Enssat Lannion, Lannion, France; the University of Bologna, Bologna, Italy; and Mid University Sweden, Sundsvall, Sweden; where is a Full Visiting Professor with the Department of Electrical Engineering. He has authored more than 300 papers in international journals and conferences. His current research interests include smart sensing, low-power machine learning, wireless sensor networks, wearable devices, energy harvesting, low-power management techniques, and extension of the lifetime of battery-operated devices. He is an ACM Member. Some of his publications were awarded Best Paper awards at several IEEE conferences. He also received awards for industrial projects or patents.
\end{IEEEbiography}

\begin{IEEEbiography}[{\includegraphics[width=1in,height=1.25in,clip,keepaspectratio]{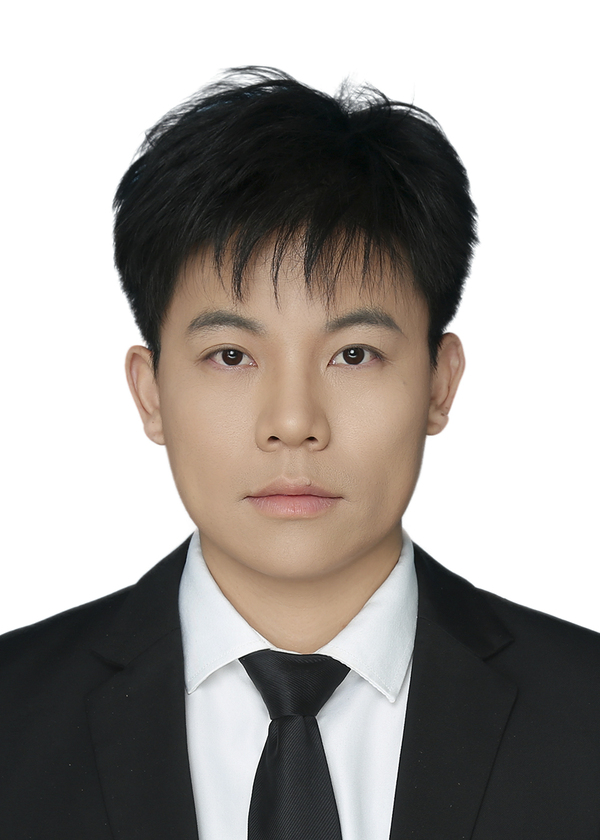}}]{Yulun Zhang} received a B.E. degree from the School of Electronic Engineering, Xidian University, China, in 2013, an M.E. degree from the Department of Automation, Tsinghua University, China, in 2017, and a Ph.D. degree from the Department of ECE, Northeastern University, USA, in 2021. He is an associate professor at Shanghai Jiao Tong University, Shanghai, China. He was a postdoctoral researcher at the Computer Vision Lab, ETH Zürich, Switzerland. His research interests include image/video restoration and synthesis, biomedical image analysis, model compression, multimodal computing, large language models, and computational imaging. He is/was an Area Chair for CVPR, ICCV, ECCV, NeurIPS, ICML, ICLR, IJCAI, ACM MM, and a Senior Program Committee (SPC) member for IJCAI and AAAI.
\end{IEEEbiography}

\clearpage

\end{document}